\documentclass[10pt]{article}

\usepackage{graphicx}
\usepackage{amsmath}
\usepackage{amsfonts}
\usepackage{amssymb}
\usepackage{amsthm}
\usepackage{wrapfig}
\usepackage{authblk}
\usepackage{url}

\usepackage{caption}
\usepackage{subcaption}
\usepackage{courier}

\usepackage[T1]{fontenc}
\usepackage{bigfoot} 
\usepackage[numbered,framed]{matlab-prettifier}

\usepackage{filecontents}

\usepackage[utf8]{inputenc}
\usepackage[english]{babel}

\usepackage{xcolor}

\usepackage{hyperref}

\newtheorem{theorem}{Theorem}[section]

\newtheorem{definition}[theorem]{Definition}

  {\begin{list}{}%
          {\setlength{\leftmargin}{#1}}%
          \item[]%
  }
  {\end{list}}

\theoremstyle{definition}

\theoremstyle{remark}

\begin{document}
 \title{The Automatic Quasi-clique Merger algorithm (AQCM)}

\author{Scott Payne} \affil{West Virginia University}
\author{Edgar Fuller} \affil{Florida International University}
\author{George Spirou} \affil{University of South Florida}
\author{Cun-Quan Zhang} \affil{West Virginia University}

\maketitle

\begin{abstract}
The Automatic Quasi-clique Merger algorithm is a new algorithm adapted from early work published under the name QCM (quasi-clique merger) \cite{Ou2007,Ou2006,Zhao2011,Qi2014}. The AQCM algorithm performs hierarchical clustering in any data set for which there is an associated similarity measure quantifying the similarity of any data i and data j. Importantly, the method exhibits two valuable performance properties: 1) the ability to automatically return either a larger or smaller number of clusters depending on the inherent properties of the data rather than on a parameter 2) the ability to return a very large number of relatively small clusters automatically when such clusters are reasonably well defined in a data set. In this work we present the general idea of a quasi-clique agglomerative approach, provide the full details of the mathematical steps of the AQCM algorithm, and explain some of the motivation behind the new methodology. The main achievement of the new methodology is that the agglomerative process now unfolds adaptively according to the inherent structure unique to a given data set, and this happens without the time-costly parameter adjustment that drove the previous QCM algorithm. For this reason we call the new algorithm \emph{automatic}. We provide a demonstration of the algorithm's performance at the task of community detection in a social media network of 22,900 nodes.
\end{abstract}

\noindent{\bf Funding.} This work was supported in part by NIH Grant R01 DC015901 and NSF Grant DMS-1700218.\\

\section{Introduction}
\label{Intro}

The task of finding a ``natural'' or automatically determined number of clusters in an arbitrary data set has long been known impossible in terms of provable optimality \cite{Jain2010}. That is, it is known that there cannot be one mathematical definition of optimality and one algorithm that finds the true optimal solution for all clustering tasks and data sets. Nevertheless, in both science and industrial data mining applications it is often necessary to discover, within a data set, clusters that have some measurable degree of quality in terms of their distinctness and the way they characterize the data according to some aspect prescribed by the application \cite{Xu2010}. Many methods have been studied over decades and our work does not aim to present a survey so in our treatment of the background of this field we will mention only a few examples of the main approaches to automatic clustering.

\subsection{AQCM}
\label{aqcm_intro}
The automatic quasi-clique merger (AQCM) algorithm is a rework of the earlier quasi-clique merger (QCM) algorithm \cite{Ou2007,Ou2006,Zhao2011,Qi2014}.
An explanation of AQCM subroutines and illustration of their implementation is given in section \ref{descr_subs}. Here we begin with a basic description. AQCM is an agglomerative hierarchical clustering approach that takes as input an $n \times n $ similarity matrix $S$ (or equivalently, a list of $ {n \choose 2}$ values representing the similarities for each pair of data points \footnote{the software we have developed requires the full $ n \times n$ similarity matrix for the purpose of faster matrix computations in various sections of the algorithm, however the mathematical procedures of the algorithm require simply the $ {n \choose 2}$ similarity values}). In this work the definition of a similarity matrix $S$ (or the $ {n \choose 2}$ values) is understood to be any function $ S : X \times X \rightarrow \mathbb{R}^+$ where $X$ is the data set and $ \mathbb{R}^+$ is the non-negative real numbers, and $ S[i,j] = S[j,i]$ for all $i,j$. Additional assumptions on the nature of similarity data input to AQCM are discussed in 
Section~\ref{clusttask}.
 For a detailed study of similarity measures in clustering see \cite{SimComp2015}. The objective of AQCM's algorithm design is to allow clusters to grow locally from well chosen pairs of data points (called seeds) known to be members of the same cluster.
Seed selection and cluster growth adapt to local properties in the similarity data so that data points may be in more than one seed pair when appropriate and clusters may grow to a size suitable for preserving relative local similarity density. Hence, the algorithm supports multi-membership (sometimes referred to as fuzzy clustering) and clustering where cluster sizes vary widely (some small clusters and some large clusters) \footnote{the algorithm discovers clusters with such properties when those properties are inherent in the data, the algorithm does not force such properties onto the clustering output}. The process repeats iteratively, each time treating the set of grown clusters as a smaller data set representing a higher level in the hierarchical scheme. The algorithm is finished running when there is only one cluster encompassing the data. While agglomerative methods are common in hierarchical clustering, the unique design of the subroutines in AQCM drive its adaptive properties and relatively efficient running time. These characteristics and the utility we have observed in our experimentation lend evidence that the algorithm can prove an important tool for obtaining automatic clustering in large and complex scientific data sets.

The AQCM method not only produces a generalized hierarchy tree (see section \ref{out_struct} for definition), but also a clustering output automatically determined by an optimization method drawing on the properties of the hierarchy tree as a graph (network). The optimization strategy was originally presented in \cite{Qi2014} where \emph{density drop} was defined as the drop in local cluster density moving up successive levels along the branches of the hierarchy tree. In that work~\cite{Qi2014}, Qi, et al., proposed that an optimal clustering can be defined as the ``cut'' \footnote{the graph theoretic term ``cut'' is defined in the section notations and definitions} across the tree featuring the most dramatic drop in cluster density. The advantage of the strategy is that tools from network science, such as max flow/min cut, provide optimization solutions for such a cut, and the resulting cut is not restricted to one level of the tree, thus allowing it to adapt to differences in local topological properties of the data along different branches of the tree. New in the work presented here is that we employ a recently proven mathematical optimization \cite{Payne2019,Payne2020edgecuts} for the selection of such a cut. The new method was designed to increase the flexibility to discover a potentially large number of clusters in a data set. We have found the method to perform as intended, and it appears to be robust in its ability to return either a small or large number depending on the inherent structure of the input data.

To speak to the intended use of the AQCM approach it is helpful to mention common uses of hierarchical clustering. In some hierarchical methods, the generated hierarchy tree is intended simply as a source for several ``resolution levels'' of clustering from which the user should choose one clustering output by some target optimization function. The term resolution refers to the granularity of clusters at a given hierarchical level, many small clusters being fine grain (i.e. high resolution) fewer larger clusters at an upper hierarchy being coarse grain (i.e.  lower resolution). Basic divisive (top-down), binary hierarchical methods are often used as such, while the Louvain method of community detection \cite{Blondel2008} is a more complex example of that use. In contrast, other methods aim to yield a hierarchy tree which characterizes complex systems of topological hierarchies inherent in a data set. Such hierarchical use is often described as ``multi-resolution'', where the purpose is to study the relationships between levels and the way they might relate to other metadata in a scientifically relevant way. The method of Jeub et al \cite{Jeub2018} is an effective example of this type of use in network data analysis.

In the context of the strategies mentioned above we can suggest the role for the AQCM algorithm is similar to the latter example while having the benefit of an optimal clustering as in the former. Although one could potentially use the automatically chosen clustering within a larger industrial data mining pipeline, AQCM might be best used in exploratory scientific data mining where not much is known about a given data set but complex hidden patterns of clustering are thought to exist. By visually and/or computationally interpreting the hierarchy tree output by AQCM a user could discover various clustering patterns that may correlate with other known metadata. Such an investigative approach often plays an important role in biology for example as described in \cite{Xu2010}. We have found AQCM to perform very well in this role in a number of data types and scientific investigations. In section \ref{facebook_section} we provide a demonstration of the algorithm's performance at the task of community detection in a social media network of 22,900 nodes. The algorithm's ability to respond automatically to inherent local properties in a data set while at the same time maintaining a relative short runtime make it a powerful tool in applications where data size and complexity both present large challenges.

\subsubsection{Software}
Implementations of the main algorithms and post processing subroutines featured in this work are provided here \cite{AQCMcode}. We chose MATLAB as the implementation platform due to its optimized performance for matrix operations. Many of the mathematical procedures in the subroutines of AQCM can be efficiently expressed as matrix operations. We found that MATLAB's combination of runtime speed and ease of dynamic memory implementation made it an ideal choice for the scalable scientific work for which we envision that AQCM is best suited.

\subsection{Automatic clustering, a short background}
\label{auto_clust}
The naive approach for automatic determination of clusters would be to repeatedly run the k-means algorithm many times over many different choices of cluster number (k) and then choose the clustering output with the highest k-means quality function score. In many applications, however, this strategy is unfeasible for a myriad of reasons, the most obvious being time constraints and the fact that we must presume that the data is best represented as a set of n-dimensional points (many data sets are not represented as such). But moreover, the situation becomes untractable in cases where a data set is known to present some large but unknown number of clusters, rendering futile the job of choosing the range of k values.

A pivotal work in the development of automatic clustering techniques was the \emph{affinity propagation algorithm} \cite{Frey2007}, or AP for short. The algorithm provides an excellent example of modern strategies that harness the power of Bayesian-like networks and allow a dynamic process to gravitate to an equilibrium.  Specifically, AP relies on an adapted version of \emph{belief propagation} \cite{Pearl1982} that Frey/Dueck refer to as ``loopy belief''. AP is in a class of strategies that rely on the notion that a data cluster has a center, often referred to as a \emph{medoid}. Also key to AP is that the input to the algorithm is not the data itself but rather a \emph{similarity matrix} which must be an $ n \times n$ real valued matrix where the ij entry scores how similar are data i and data j.

Another class of clustering strategies relies only indirectly on the notion of a medoid and treats clustering as the problem of fitting a finite mixture model to the input data set. Many algorithms exist but a particularly effective method found here \cite{Figueiredo2002} builds upon various strategies that optimize log-likelihood and can be adapted to various statistical models (Gaussian being the default).

A final example category of clustering strategies is found in the graph theoretic task of \emph{community detection}. Any similarity matrix representing a data set can be thought of as a weighted graph (network) where each vertex (node) represents a data point. Communities are groups of vertices within which the edge weights (similarities) are relatively high and edge weights between communities tend to be lower. So community detection is equivalent to clustering when an appropriate similarity matrix exists. There is a large literature on community detection. The main early example is found in the work of Girvan/Newman \cite{Girvan2002}. The methods proposed by Girvan/Newman are driven by the optimization of a target function called \emph{modularity}. A key advancement in modularity optimization techniques was the algorithm called \emph{Louvain method} \cite{Blondel2008} which allowed for very large data sets to be processed in feasible time on an appropriate system. A hierarchical technique that generalizes the Louvain method is found here \cite{Jeub2018} and a very recent modularity generalization algorithm is found here \cite{Trag2019}. Importantly, that work (the latter) provides theorems proving that the quality of the clusters (according to a user chosen target function and resolution parameter) is optimized by further iteration.

As mentioned above, the review of methods provided here is only intended to illustrate some of the existing strategies. We have focused on strategies that aim to produce clusters where the number suits the unique properties of the input data set in some ``natural'' way without the requirement that the user have foreknowledge of the ``correct'' number of clusters to find. It is important to take a moment to note, however, that in many cases a data set may have multiple equally valid solutions to the clustering task or none (for discussions see \cite{Jain2005} and \cite{Peel2017}). Furthermore some clustering tasks, such as those designed to divide data for distributed computing, are such that the user \emph{knows} the desired number of clusters ahead of time. Hence, in the next section we clarify the assumptions we make in this work regarding the nature of clustering tasks we would aim to perform and for which the algorithms discussed above are intended.

\subsection{The clustering task}
\label{clusttask}
In this work we assume that a given clustering task on a data set has the following characteristics:
\begin{enumerate}
	\item The data set has distinct structures in that there are subsets of the data within which the points tend to be more similar to each other and less similar to points in other subsets. The difficulty of such a clustering task would be determined by the range of similarities inside subsets and the range of similarities between subsets. The more that the between range overlaps the inside range, the more difficult it is to determine whether a pair of points are in the same cluster.
	\item If the data are points in n-dimensional space we assume that the points are distributed such that there are some number of regions of higher point density separated by regions of lower density. The difficulty of such a clustering task would be determined by the distinctness of the dense regions against the space between regions and also the shape of the regions (which need not be symmetrical or spherical). For example the moon and sun problem and the target problem are construction methods that create data sets which will confound most modern clustering algorithms.
\end{enumerate}

\subsection{Notation and definitions}
\label{notation}

\subsubsection{AQCM and graph theory}
\label{aqcm_n_graphs}
Many of the subroutines in the AQCM algorithm are adaptations or redesigns of subroutines from the QCM algorithm, and the general structure of each iteration of AQCM is the same as in QCM. Above we have described AQCM as a clustering algorithm that takes as input an $ n \times n$ similarity matrix $S$, a common type of input structure in data mining \cite{SimComp2015} (see section \ref{aqcm_intro} for precise definition). It is important to note that the original publications of the QCM algorithm \cite{Ou2007,Ou2006,Zhao2011,Qi2014} presented the quasi-clique merger method as an algorithm for performing hierarchical ``community detection'', a common data mining task in graph (network) data analysis. In that work, the input to QCM was a weighted graph (weighted network) and a community was defined as an ``edge dense'' set of vertices (nodes) where edge weights tend to be higher, edges between communities were expected to be more sparsely distributed and/or lower weight. The objective of the QCM algorithm and the type of input graph for which it was designed are equivalent to the idea of clustering in similarity data as commonly discussed in data mining today \cite{SimComp2015}. Thus far we have introduced the idea of AQCM using the language of similarity-based clustering due to its useful and accurate level of generality in framing the tasks for which AQCM may be used. But relevant to note is that the QCM algorithm itself is not able to operate on an unwieghted graph, that is, the input must have weights associated with the edges which serve the same role as that of a similarity function. Hence, the description of the quasi-clique merger process in general might be most accurately described as a process that operates on similarity data.

On the other hand, while the mathematical steps in AQCM (and QCM) can be accurately described using the abstract notions of data points and similarity, many of the steps can actually be more easily discussed using their equivalent formulations in the language of graph theory. The graph theoretic concept of an \emph{edge}, sometimes also referred to as a \emph{link}, is a linguistically simple way to refer to the relationship between two data points, which in graph theoretic terms would be referred to as \emph{vertices}, or sometimes \emph{nodes}. Therefore, in the work presented here we may at times describe processes in terms of vertices and edges instead of data points and similarities, and input to AQCM may at times be described as a weighted graph instead of similarity values or a similarity matrix.

\subsubsection{Graph theoretic terminology}
\label{graph_terms}
For graph theoretic terminology and notation we follow \cite{Bondy2008,West2001,Diestel2017} which are standard. A graph $ G(V,E)$, or $G$ for shorthand, is defined by a vertex set $V$ (sometimes notated $ V(G) $ ) and an edge set $E$ (sometimes notated $ E(G)$ ) where an edge $e \in E $ is a connection between some vertex $ u \in V$ and some other vertex $v \in V$ so that $e$ might also be notated $ uv$. It is common to notate the size of the vertex set $|V| = n $ and the size of the edge set $|E| = m $. If $G$ is a weighted graph then $ w : E \rightarrow \mathbb{R}$ is the function assigning weights to edges and the weight of edge $ e$ is notated $ w(e)$. Some authors assume that for a weighted graph all weights are non-zero or equivalently that $ w(uv) = 0$ implies $ uv \notin E$. In this work we will address such cases as necessary. If $G$ is a directed graph (digraph) then each edge $ uv$ has the direction $ u$ to $v$ where $v$ is referred to as the \emph{head} of the edge and $u$ is the \emph{tail}, and between two vertices $ u$ and $ v$ both the directed edges $ uv$ and $ vu$ might exist. Some authors use the term \emph{arc} for a directed edge however here we will simply use the term edge and when the edge has a direction we will make that clear in context. Some additional technical graph-theoretic terminology is provided in section \ref{app_graph}.

As discussed in section \ref{aqcm_n_graphs}, parts of AQCM are sometimes more easily described using a graph theoretic language. Here we discuss notation. Let $X$ be a data set and $S$ be a similarity function $ S : X \times X \rightarrow \mathbb{R}^+$ as defined in section \ref{aqcm_intro}, then $S$ is equivalent to a weighted graph $ G(V,E)$ with weight function $ w : E \rightarrow \mathbb{R}$ where vertex $v_i \in V$ is associated with data point $ x_i \in X$ and for a pair of vertices $ v_i , v_j$ the edge weight $ w(v_iv_j)$ is equal to the similarity value $ S(x_i,x_j)$. Clearly the discrimination between $ w(v_iv_j)$ and $ S(x_i,x_j)$ is purely semantic, so henceforth, when we adopt a graph theoretic language of algorithm description we will use the notation $ S$ and refer to it as a weighted graph and refer to $ S(x_i,x_j)$ as the edge weight for the edge $ x_ix_j$.

\subsubsection{Input and output structures of AQCM}
\label{out_struct}
The input to AQCM is an $n \times n $ similarity matrix $S$ (or equivalently, a list of $ {n \choose 2}$ values representing the similarities for each pair of data points) where $n$ is the number of data points in some data set $X$. The definition we adopt for the terminology \emph{similarity} was given in section \ref{aqcm_intro} and further assumptions on the nature of similarity data input to AQCM are discussed in section \ref{clusttask}.

The output of a hierarchical clustering algorithm is typically represented by a dendrogram which is simply a diagram of a hierarchy tree. A precise graph theoretic definition of the term \emph{tree} is given in appendix \ref{app_graph}. A graph that is a tree is typically given the notation $T$ instead of $G$. The \emph{rooted tree} (see appendix \ref{app_graph} for precise definition) is the proper graph theoretic object that characterizes the concept of a hierarchy tree. The root represents the data as one set, the leaves represent the data as individuals, and nodes along a path from the root to a leaf represent cluster assignment for that leaf at various hierarchical levels. In a dendrogram, the depiction is typically oriented with the root at the top of the diagram and the leaves (representing individual data points) at the bottom. In data science, hierarchical relationships are often described using the terminology \emph{parent} and \emph{child}. In a hierarchy tree $T$, a parent node $v \in V(T)$ represents a cluster at some level ($v$ is not a leaf node). For a node $u \in V(T)$ such that there is an edge $vu \in E(T)$ we say $u$ is a child node of the parent node $v$. The node $u$ represents a sub-cluster (of the parent cluster represented by $v$) at the next level down in the hierarchical scheme (or an individual data point if the parent is a level 1 cluster).

 As mentioned in section \ref{aqcm_intro}, the hierarchical information output by AQCM supports multimembership. That is, when appropriate, a data point (or cluster at some level in the hierarchy) could be a member of more than one cluster. In parent/child terminology we say a multimember child node has more than one parent node. Hence the hierarchy ``tree'' output by AQCM may not be a true tree according to the graph theoretic definition \emph{tree} (see appendix \ref{app_graph}). For simplicity of language in this work we will nonetheless refer to the hierarchical output produced by AQCM as a tree since we have clarified here the context and the idea of a generalized hierarchy tree (mentioned in section \ref{aqcm_intro}) that allows for multimembership.

Based on the above we should mention that, in our method, the automatically chosen clustering selected from the hierarchy tree is not necessarily a \emph{partition} of the data set (although it could be) but is more generally a \emph{family} of subsets of the data notated $ \mathcal{C} = \lbrace C_1,...,C_k \rbrace$ where $C_i \subset X$ is the $i$th cluster.  Also, the clustering output $ \mathcal{C}$ may or may not form a \emph{covering} of the data set as some outlier points might not be assigned to any cluster. We will refer to such unassigned points as \emph{singletons} or simply \emph{unclustered data}. Examples are seen in section \ref{example_data_1}.

A final aspect of our method that we will clarify here is the relationship between automatic cluster selection and the graph theoretic notion of an optimal \emph{edge cut}. The terminology \emph{edge cut} is defined precisely in appendix \ref{app_graph}, but here we can describe its relevance in a hierarchy tree as a rooted tree \footnote{For simplicity of explanation we describe the case of a tree with no multimembership, however the relationship exists similarly in trees featuring multimembership.}. Consider that every cluster $C$ defined within the structure of hierarchy tree $T$ is represented by a child node of some parent node (possibly the root). Hence, for any clustering output $ \mathcal{C} = \lbrace C_1,...,C_k \rbrace$ selected from the tree $T$, there is an associated set of directed edges $ H = \lbrace e_1,...,e_k \rbrace \subset E(T)$ where $ e_i = v_iu_i$ is associated with cluster $C_i$ through its representation by child node $u_i$. The set of edges $H$ is called an \emph{edge cut} because its removal separates the tree $T$ into components. The selection of optimal edge cuts is a common topic in the theory of graphs and networks. Hence the selection of an optimal clustering $ \mathcal{C} = \lbrace C_1,...,C_k \rbrace$ from hierarchy tree $T$ may be approached by the selection of an optimal edge cut $ H = \lbrace e_1,...,e_k \rbrace \subset E(T)$ provided that a suitable method exists for encoding, into the edges of $T$, information about the various clustering outputs available within $T$. We present our implementation of this approach in section \ref{clust_sel}.

%

\section{An Illustrated Introduction to AQCM}
\label{example_data_1}
\begin{figure}[!htb]
\centering
\includegraphics[scale=0.35,trim=0 0 0 0]{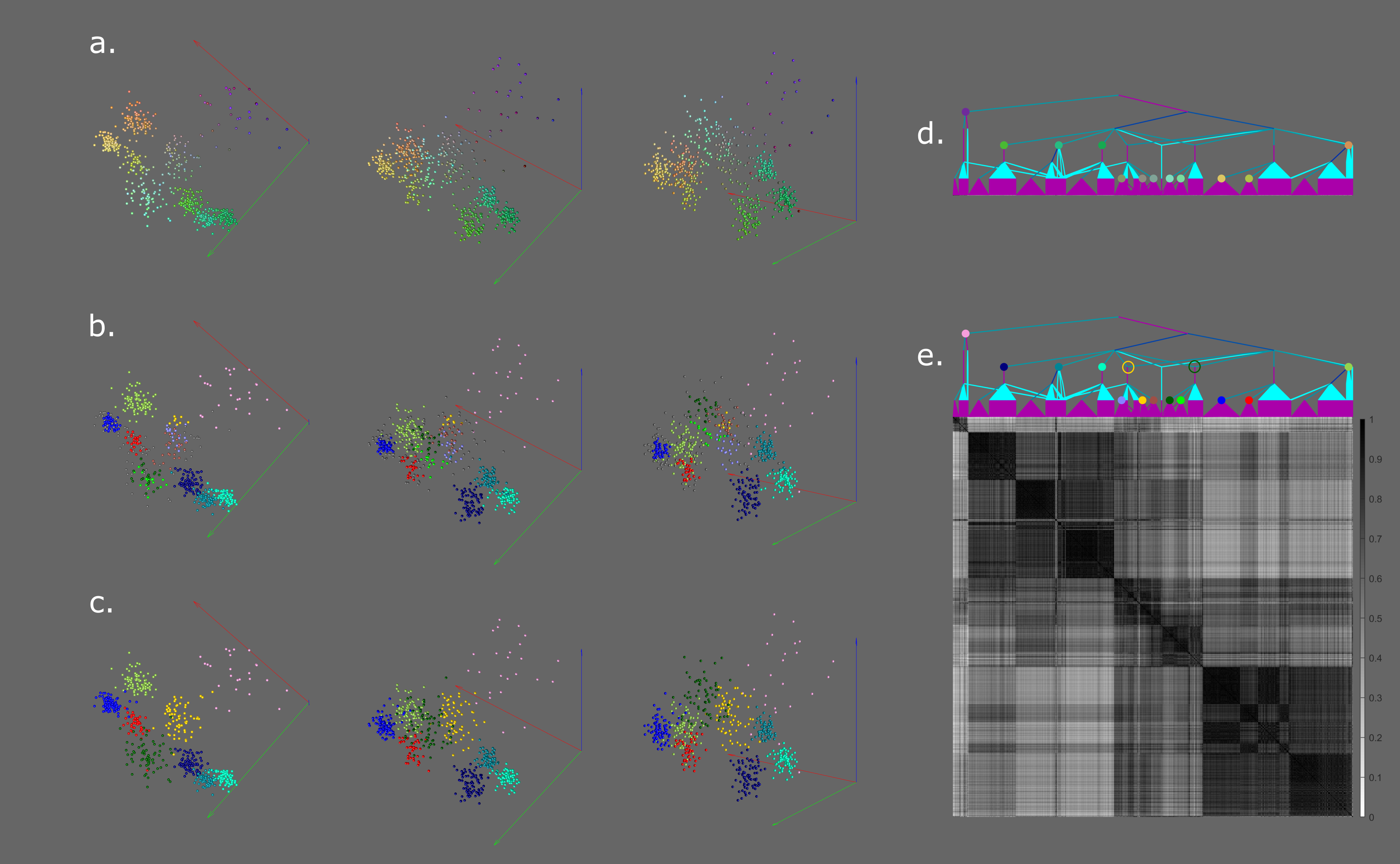}
\captionsetup{font=footnotesize}
\caption{\textbf{a.} 3-dimensional (RGB) data seen in three views, RGB axes are indicated by their color, each data point is colored by it's RGB value, each axis range is [0,1]    \textbf{b.} three views match fig 1a, the data are colored to distinguish the clusters chosen by AQCM and match the indicator colors in the dendrogram seen in fig 1e, data colored gray are ``unclustered/outliers''    \textbf{c.} as in fig 1b but here the colors distinguish the clusters obtained using a Gaussian finite mixture model fitting algorithm described in \cite{Figueiredo2002}, we may compare and contrast this clustering with that obtained by AQCM    \textbf{d.} the dendrogram output by the AQCM algorithm, the layout of the dendrogram was computed using Matlab's ``layered layout'' implementation for directed graphs, the colors of the lines indicate the similarity density drop between hierarchical levels according to the function described in section \ref{clust_sel} with light blue being more extreme dark blue moderate and purple least extreme, the colored dots indicate the automatically chosen clustering and their colors are the mean color per cluster    \textbf{e.} the dendrogram as described in fig 1d except that here the cluster indicators are colored to match the illustration in fig 1b, below the dendrogram the similarity data is displayed as organized by the dendrogram (this type of display is often referred to as a ``heat map'') and the scale is given at the right, the yellow and green open circles show two alternate clusters represented in the AQCM output and these are essentially the same as the yellow and green clusters seen in fig 1c (only differing by a few data)}
\label{example_1_fig}
\end{figure}

In this section we demonstrate the use of the AQCM algorithm to perform unsupervised cluster analysis on a basic abstract data set. The set was designed to exemplify the type of clustering tasks described in section \ref{clusttask}. 3D renderings of the data points are seen in figures \ref{example_1_fig}a,b,c. By mapping the data into the cube on the interval [0,1], it can be interpreted as RGB color data (as illustrated in figure \ref{example_1_fig}a). The colors allow us to intuitively understand the inherent hierarchies and relationships between clusters. 

\subsection{The data set}
\label{data_set_1}
The  construction of the data set was as follows: we chose 9 3-dimensional points randomly and centered Gaussian distributions at those points. Covariances were chosen so that samples from the mixture model would generate point clouds with reasonably distinct boundaries as discussed in section \ref{clusttask}. Samples were drawn of random sizes from each Gaussian in the mixture model so as to generate planted clusters of different sizes (in terms of number of data points). There are 595 data points in total. The construction method is standard for synthetically generating point clouds with clustering properties similar to those often seen in experimentally collected scientific data.

\subsection{Benchmark algorithm}
\label{benchmark_1}
For a benchmark/comparison we used the finite mixture model fitting algorithm of Figueiredo et al described in \cite{Figueiredo2002} to automatically determine the cluster assignments illustrated by color in figure \ref{example_1_fig}c. The algorithm discovers 9 clusters whose centers and geometry match the underlying model used to generate the data, so we may think of this clustering as ``correct'' in that sense. We have tested this algorithm extensively in similar data sets and have found that it appears to return an accurate description of the planted clusters as long as the number of clusters is not too large and the differences in their relative geometries not too extreme. The algorithm is essentially parameter free for simple data sets with relatively few clusters since the required input range of cluster numbers to try can be fixed to an interval such as [1,25] (that is what we used here). We have found that when the properties of the planted clusters become more extreme (in terms of number, size and geometry) the algorithm tends to produce less accurate results even if a good parameter range is used. We should also note that we have not benchmarked this algorithm's performance in high dimensional data , so we cannot say whether it is feasible for such data in terms of accuracy or run time.

\subsection{AQCM algorithm output}
\label{AQCM_example_1}
Illustrations of AQCM output are seen in figures \ref{example_1_fig}b,d,e. Figures \ref{example_1_fig}b,c are color-coded diagrams of the clustering outputs of AQCM and the benchmark algorithm respectively \footnote{we have ``aligned'' the colors in the illustration for ease of comparison between the two clustering outputs}. 

As AQCM requires a similarity matrix, we defined similarity as the negative of Euclidean distance shifted into the range 0 to 1. The similarity data is seen in figure \ref{example_1_fig}e as indexed by the hierarchy tree output by AQCM (seen above the matrix).

We begin by observing that the hierarchy tree output by AQCM has, on a high level, three main clusters (figure \ref{example_1_fig}d shows these): the purple dot indicates the average color of the cluster on the left branch while the right branch splits into two subclusters. Colored dots of further subclusters indicate that the two containing clusters represent a green group and a yellow/orange group with brown/tan/bluegrey clusters having multimembership of both green and yellow/orange. In the visual display of the color data seen in figure \ref{example_1_fig}a it is certainly clear that splitting the data in two between purple and non-purple data  is a valid high level distinction and that the non-purple data can be described as yellow/orange and green with brown/tan/bluegrey in the middle. Hence the general structure of the hierarchy tree captures the inherent nature of the data well. 

Using figures \ref{example_1_fig}b,c,e we can examine the clustering output selected by edge-cut cluster detection. The colors in figures \ref{example_1_fig}b,e illustrate that clustering. We have synchronized our labelling colors for figures \ref{example_1_fig}b,c so that the AQCM clustering may be directly compared with the benchmark algorithm clustering described above. The yellow/orange group and the green group each contain clusters that match the benchmark algorithm almost perfectly. The differences here are on account that AQCM leaves some data unclustered (see the red labelled and blue labelled clusters). This aspect of AQCM is a natural occurrence when a set of densely clustered points has outliers, AQCM may prefer the denser subset since it optimizes for cluster similarity density. 

The main difference between the chosen clusters of edge-cut detection and the benchmark algorithm are in the treatment of the green labelled and yellow labelled clusters seen in figure \ref{example_1_fig}c. As illustrated by the open circles in figure \ref{example_1_fig}e, AQCM captures these same clusters at a higher hierarchy. Examination of the color data itself in figure \ref{example_1_fig}a would seem to indicate that the distinctions made by AQCM to split these two clusters further could also be considered valid since they each respectively contain locally dense similarity subsets as seen in figure \ref{example_1_fig}e. We should remark here that this difference between AQCM output and the benchmark algorithm demonstrates the different interpretations of data clusters given when one algorithm determines structure based on distance and another attempts to fit a Gaussian distribution. Hence the differences between the two outputs are inherent in the different definitions of the feature space of the data. Overall, the color data example demonstrates AQCM’s ability to function well even within a feature space that presents challenges in terms of fuzzy clusters and regions of varying data density. One might expect that if the similarity function on the data were able to reflect the planted Gaussian clusters even more clearly then the clustering detected by the edge-cut algorithm might even more precisely match that of the benchmark algorithm. In general, the description of the data set provided by the hierarchy tree output does provide a robust multi-level interpretation of the panted structures as we might naturally see them with our human ability to perceive groups and dense point clouds.

%

\section{AQCM Algorithm and Subroutines}
\label{descr_subs}

In section \ref{aqcm_algo_struct} we provide a basic description of each of the subroutines of AQCM so as to facilitate an easier understanding of their objectives and mathematical strategies. An illustration of the process is seen in figure \ref{subs_fig}. Full mathematical details are given in technical notation in section \ref{sub_det}. As discussed in section \ref{aqcm_n_graphs}, parts of AQCM are sometimes more easily described using a graph theoretic language. Here we adopt that style to describe the subroutines and their unfolding. See section \ref{graph_terms} for clarification of the relationship between our notation and graph theoretic language. 

\subsection{AQCM algorithm structure}
\label{aqcm_algo_struct}

AQCM has four subroutines: \emph{seed selection}, \emph{growth step}, \emph{adjustment step}, \emph{contraction step}. The AQCM algorithm iterates by running the subroutines in that order, each outputting objects for input to the next subroutine. The contraction step creates the input graph for the seed selection of the next iteration.

\begin{figure}[!htb]
\centering
\includegraphics[scale=0.58,trim=25 40 0 0]{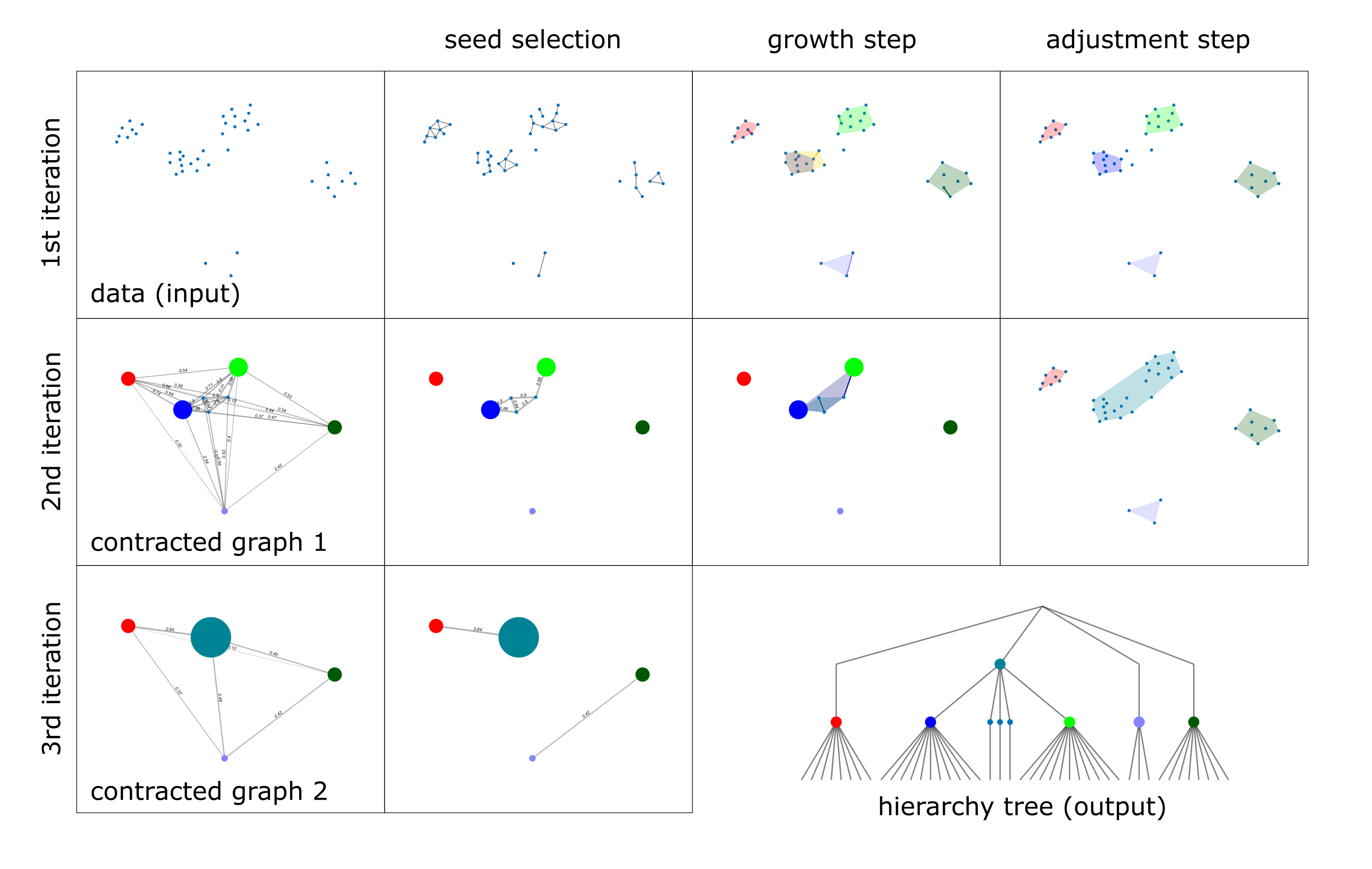}
\captionsetup{font=footnotesize}
\caption{\textbf{a.} An illustration of each subroutine's output as AQCM runs on a simple data set with planted clusters of 2D points (seen in the top left box). The similarity function we used was the ``shifted negative of distance'' as described in section \ref{AQCM_example_1}. Each row of the diagram shows an iteration (or hierarchical level) running through the processes of seed selection, growth, and adjustment. For each subroutine, its column in the diagram shows the relationship between the ways the algorithm views the data at different hierarchical levels. In column 1, iterations 2 and 3 show the \emph{contracted graph} that will be input to seed selection, dot size reflects cluster size and their color is associated with steps in the previous iteration, the edge weights are seen to two decimal places. In illustrations of the growth step, the seeds from which the illustrated clusters grew are seen as colored edges. For this example data set, the 3rd iteration growth step resulted in one cluster covering the data, so the algorithm was finished at that step. Bottom right shows the dendrogram of the hierarchical clustering structure determined by the states of clustering at the end of each iteration, colored node indicators show the relations to the illustrated subroutines.}
\label{subs_fig}
\end{figure}

\begin{tabbing}
\noindent \= Input: \hspace{3 pt} \= $ S : X \times X \rightarrow \mathbb{R}^+$ a similarity function on a data set $X$, equivalently, a weighted graph \\ 
\> \> on a vertex set $X$ where $ S(x_i,x_j)$ is the edge weight between vertices $ x_i$ and $x_j $, in the \\
\> \> first iteration of AQCM this is the current graph, though in the notation below we use \\
\> \> the letter $V$ instead of $X$ in order to generalize the language for all iterations.\\
\noindent \= Output: \= $T$ a generalized hierarchy tree, (see section \ref{out_struct} for terminology)\\
\vspace{0.2 in}
\indent \= \qquad \= \qquad \= \qquad \= \qquad \= \qquad \= \qquad \= \\
\> \textbf{seed selection:} \> \> \> \> \> Select a subset $ E_{seeds} \subset E$  ($E$ is the edge set of the current graph on a \\
\> \> \> \> \> \> vertex set $V$) such that for $ e = v_iv_j \in E_{seeds}$, $ v_i$ and $ v_j$ are likely to be \\
\> \> \> \> \> \> in the same cluster. This likelihood is ensured by estimations of locally \\
\> \> \> \> \> \> close vertices made at each vertex: a seed edge $ e = v_iv_j \in E_{seeds}$ is an \\
\> \> \> \> \> \> edge such that in the estimations at vertex $ v_i$ and vertex $ v_j$, each \\
\> \> \> \> \> \> designated the other as a locally close vertex. \\
\\
\> \textbf{growth step:} \> \> \> \> \> Grow a set of clusters $ \mathcal{C} = \lbrace C_1, \ldots ,C_k \rbrace$ by optimally adding vertices of \\
\> \> \> \> \> \> $V$ to individual seed edges of $E_{seeds}$. Growth is controlled locally in that \\
\> \> \> \> \> \> for a given cluster $ C$ growing from a seed $e \in E_{seeds}$, $C$ stops growing \\
\> \> \> \> \> \> when its edge weight density falls below a threshold developed dynamically\\
\> \> \> \> \> \> from the cluster's growth properties. The process unfolds linearly starting \\
\> \> \> \> \> \> with the highest weight seed edge in $E_{seeds}$. As a cluster $C$ grows covering \\
\> \> \> \> \> \> other seed edges, those edges are removed from $E_{seeds}$ eliminating unnecessary \\
\> \> \> \> \> \> processing. \\
\\
\> \textbf{adjustment step:} \> \> \> \> \> Adjust the clustering $ \mathcal{C} = \lbrace C_1, \ldots ,C_k \rbrace$ when there are subsets of $ \mathcal{C}$ with \\
\> \> \> \> \> \> relatively large vertex overlap. Such subsets are possible since the \\ 
\> \> \> \> \> \> seed/growth process may approximate a potential cluster several slightly \\
\> \> \> \> \> \> different ways. In order to ensure that the most locally dense possible \\
\> \> \> \> \> \> cluster is not overtaken by a less optimally dense approximation, the first \\
\> \> \> \> \> \> iteration uses a different adjustment method than subsequent iterations.\\
\> \> \> \> \> \> \textbf{First Iteration:} working through $\lbrace C_1, \ldots ,C_k \rbrace$ in order prioritizing \\
\> \> \> \> \> \> whichever cluster $C$ is most edgeweight dense, remove any other cluster\\
\> \> \> \> \> \> that overlaps $C$ too much. Processing continues until degenerate overlaps \\
\> \> \> \> \> \> are resolved.\\
\> \> \> \> \> \> \textbf{Subsequent Iterations:} working through $\lbrace C_1, \ldots ,C_k \rbrace$ in order prioritizing \\
\> \> \> \> \> \> whichever cluster $C$ is most edgeweight dense, merge into $C$ any other \\
\> \> \> \> \> \> cluster that overlaps $C$ too much. When more than one merge is possible, \\
\> \> \> \> \> \> choose the merge resulting in the highest edgeweight density. Processing \\
\> \> \> \> \> \> continues until degenerate overlaps are resolved. Note that for iterations two\\
\> \> \> \> \> \> and higher, merge decisions are made by interpreting clusters as subsets of \\
\> \> \> \> \> \> the data set $X$ as opposed to subsets of contracted graph vertices (see \\
\> \> \> \> \> \> figure \ref{subs_fig} second row last column).\\
\\
\> \textbf{contraction step:} \> \> \> \> \> Create a weighted graph representing the similarities between each pair \\
\> \> \> \> \> \> of clusters $ C_i$ and $C_j $. As in the adjustment step, clusters are interpreted \\
\> \> \> \> \> \> as subsets of the data set $X$ as opposed to subsets of contracted graph \\
\> \> \> \> \> \> vertices. Each cluster $C_i$ is represented as a vertex $v_i$ in the new graph, \\
\> \> \> \> \> \> the edge weight between vertices $ v_i$ and $v_j $ is the mean of all edgeweights \\
\> \> \> \> \> \> of edges (of the graph on $X$) between $ C_i$ and $C_j $.  The weighted graph \\
\> \> \> \> \> \> created by the contraction step becomes the current graph which is the \\
\> \> \> \> \> \> input to the next iteration of AQCM beginning with seed selection.\\
\end{tabbing}

\subsection{Subroutine details}
\label{sub_det}
In the following descriptions of the AQCM subroutines we use a more proper graph theoretic notation style. Let $S$ be a weight function $S : E(K_n) \rightarrow \mathbb{R}^+$ where $E(K_n)$ is the edge set of a complete graph $K_n$ on $n$ vertices with vertex set $V$. We define the following.

\begin{definition}
\label{DEF: density}
For a subgraph $Q$ of $K_n$ induced by a vertex subset $ C \subset V$,
 we define the density of $Q$ by
$$\mathrm{den}(Q) ~ = ~ \frac{2 \sum_{e \in E(Q)} S(e)}{|C|(|C|-1)}$$
\end{definition}

As $Q$ is an induced subgraph of a complete graph, we have $ |E(Q) | = {|C| \choose 2}$. So it is easy to see that the function $ \mathrm{Q}$ is simply the mean edgeweight over the subgraph $Q$, or equivalently, the average similarity between pairs of data in the cluster $C$.

\begin{definition}
\label{DEF: contribution}
For a vertex $v \notin V(Q )$, we define the
contribution of $v$ to $Q$ by
$$\mathrm{cont}(v,Q) ~ = ~ \frac{\sum_{u \in V(Q )} S(uv)}{|V(Q )|}.$$
\end{definition}

The contribution $ \mathrm{cont}(v,Q)$ is the mean of edgeweights between $v$ and $Q$, and as such it is clearly the simple and computationally efficient way to quantify a vertex's candidacy for joining a growing cluster.

\subsubsection{Seed selection}
\label{seed_sel}
Seed selection is a new subroutine in AQCM not found in published versions of QCM. It's purpose is to automatically select seeds from which to grow clusters in a way that adapts to the inherent properties of the input data. The intuition driving the seed selection algorithm comes from the assumptions we often make about what distinguishes a cluster within a data set (see section \ref{clusttask}). We expect that for a data $x$ which is well within (not an outlier) a cluster $C$, the similarity values of $x$ to other data in $C$ would be in a numerical range that should be somewhat distinct from the numerical range of similarity values between $x$ and the data not in $C$. In our seed selection algorithm we harness that property by considering, for data $x$, the list of similarity values to other data sorted in decreasing order. Thus we should expect some position in that ordered list where the similarity value abruptly drops as we move from similarities measured within $C$ to those with other data outside $C$. While this property might not hold for every data $x$ in $C$, it is easy to show that our algorithm will produce at least one seed edge inside each cluster and our algorithm will not choose edges representing a pair of data each in a different cluster assuming the similarity data has reasonable clusters as described in section \ref{clusttask}.\\

\noindent \underline{\textbf{seed selection algorithm:}}
\begin{tabbing}
\noindent \= Input: \hspace{3 pt} \= $K_n$ a complete graph on $n$ vertices, and edgeweight function $S : E(K_n) \rightarrow \mathbb{R}^+$,\\ 
\> \> when AQCM is in the first iteration $K_n$ and $S$ are the inputs to AQCM, in \\
\> \> subsequent iterations $K_n$ and $S$ are objects created by the previous iteration \\
\> \> contraction step.\\
\noindent \= Output: \= the seed edge set $ E_{seeds} \subset E(K_n) = \lbrace e_1,...,e_m \rbrace$, for use in the growth step the \\
\> \> set $E_{seeds} $ should be output as an ordered set $ \lbrace e_{r_1},...,e_{r_k} \rbrace$ with $\lbrace r_1,...,r_k \rbrace \subset \lbrace 1,...,m \rbrace $ \\
\> \> and $ S(e_{r_i}) \geq S(e_{r_{i+1}})$ for all $i$.\\ 

\indent \= \qquad \= \qquad \= \qquad \= \qquad \= \qquad \= \qquad \= \\
\noindent \textbf{Step 1.}\\
\> \> \textbf{for} each vertex $ v \in V(K_n)$ \textbf{do:} \\[0.15 in]
\> \> \> order vertices decreasing by edgeweight with $v$:\\
\> \> \> $ \lbrace  u_1, \ldots , u_{n-1} \rbrace$ is the set of vertices not including $v$\\
\> \> \> $ \lbrace u_{q_1}, \ldots , u_{q_{n-1}}  \rbrace $ is an ordering with $ S(vu_{q_i}) \geq S(vu_{q_{i+1}})$ for all $i$\\[0.15 in]
\> \> \> compute a sequence of differences:\\
\> \> \> $ \lbrace a_i \rbrace _{i=1}^{n-2}  \leftarrow \lbrace S(vu_{q_i}) - S(vu_{q_{i+1}})  \rbrace _{i=1}^{n-2}$\\[0.15 in]
\> \> \> compute the median of that set (treated as a statistical sample):\\
\> \> \> $ M_a \leftarrow \mathrm{median}( \; \lbrace a_{\ell} \; | \; \ell \in \lbrace 1, \ldots ,n-2  \rbrace \; \rbrace \; )$\\[0.15 in]
\> \> \> find the first significant edgeweight drop:\\
\> \> \> $ t \leftarrow  \mathrm{min}( \; \lbrace  i \; | \; a_i \geq M_a \; \rbrace \; )$\\[0.15 in]
\> \> \> store a list of locally close vertices for $v$:\\
\> \> \> $ \mathcal{L}_v \leftarrow  \lbrace u_{q_1} , \ldots , u_{q_t}  \rbrace$\\[0.15 in]
\noindent \textbf{Step 2.}\\
\> \> select seed edge set:\\
\> \> \> $ E_{seeds} \leftarrow  \lbrace  uv \in E(K_n) \; | \; u \in \mathcal{L}_v \; \text{and} \; v \in \mathcal{L}_u \; \rbrace$\\
\noindent \textbf{Step 3.}\\
\> \> sort the seed edge set $ E_{seeds}$ decreasing order by edgeweight\\
\end{tabbing}

\subsubsection{Growth step}
\label{growth}
The growth subroutine in this work follows the growth principals introduced in previous publications of QCM with the exception of a new parameterized method we introduce via the parameter $ \tau$. The method resolves the issue that occurs when more than one vertex is a numerically valid candidate to join a growing cluster $C$. Specifically, the parameter $ \tau $ is intended as a sort of numerical padding for the case when similarity calculations may have slight differences between optional joins, but those differences may be a ``digital artifact'' of the similarity matrix calculation. That is, the optimal vertex to join may have average similarity to $C$ greater by an insignificant margin than other possible vertices. $ \tau $ allows that any vertices within an acceptable threshold of optimal may join at the same time. This new method allows for correct processing of data where symmetries exists such that some groups of data should be treated equally as opposed to being arranged in a linear order of priority. Finally, we should also mention that $ \tau$ is not intended as a tuning parameter, but rather as a user determined value. We have found that $ \tau = 0.008$ has performed well in all data sets tested for similarity functions where the range is in the interval $ [0,1]$.

\noindent \underline{\textbf{growth step algorithm:}}
\begin{tabbing}
\noindent \= Input: \hspace{3 pt} \= $ E_{seeds} = \lbrace e_{r_1},...,e_{r_k} \rbrace$ with $\lbrace r_1,...,r_k \rbrace \subset \lbrace 1,...,m \rbrace $, and $ S(e_{r_i}) \geq S(e_{r_{i+1}})$ for all $i$.\\
\> \> $K_n$ a complete graph on $n$ vertices, and edgeweight function $S : E(K_n) \rightarrow \mathbb{R}^+$,\\ 
\> \> when AQCM is in the first iteration $K_n$ and $S$ are the inputs to AQCM, in \\
\> \> subsequent iterations $K_n$ and $S$ are objects created by the previous iteration \\
\> \> contraction step. For notation below we abbreviate $V(K_n)$ as just $V$.\\
\noindent \= Output: \= $ \mathcal{C} = \lbrace C_1, \ldots ,C_k \rbrace$, a family of subsets $ C_i \subset V(K_n)$ where each $C_i$ is an approximation\\ \> \> of some cluster (in the current graph) of relatively high edgeweight density as in \\
\> \> definition \ref{DEF: density} \\

\indent \= \qquad \= \qquad \= \qquad \= \qquad \= \qquad \= \qquad \= \\
\noindent Initialize: $\mathcal{C} \leftarrow \lbrace \rbrace$\\
\> \> \hspace{7pt} $ \tau \leftarrow 0.008$\\
\noindent \textbf{Step 1.} \textbf{If} $ E_{seeds} = \lbrace \rbrace$ \textbf{exit growth step algo} , \textbf{Else do:}\\[0.15 in]
\> \> start a new cluster from the maximum weight seed:\\
\> \> $C \leftarrow \lbrace u,v \rbrace$ where $ uv=e$ is the first seed in the ordered list $ E_{seeds}$\\[0.15 in]
\> \> \textbf{Step 2.} try to grow cluster $C$: \\[0.15 in]
\> \> \> \> compute $ \alpha$ value:\\
\> \> \> \> $ \alpha \leftarrow 1 - \frac{1}{2(|C| + 1)}$\\[0.15 in]
\> \> \> \> compute $ \mathrm{den}(Q)$ where $Q$ is the subgraph of $ K_n$ induced by $C$: \\
\> \> \> \> see definition \ref{DEF: density}\\[0.15 in]
\> \> \> \> compute maximum contribution:\\
\> \> \> \> see definition \ref{DEF: contribution}\\
\> \> \> \> $ q \leftarrow \mathrm{max}( \; \lbrace \mathrm{cont}(v,Q) \; | \; v \in V \setminus C \;  \rbrace \; )$\\[0.15 in]
\> \> \> \> \textbf{If} $ q \geq \alpha * \mathrm{den}(Q)$ \textbf{do:}\\[0.15 in]
\> \> \> \> \> find eligible vertices to join $C$:\\
\> \> \> \> \> $J \leftarrow \lbrace v \in V \setminus C \; | \; \mathrm{cont}(v,Q) > q - \tau \; \rbrace $\\[0.15 in]
\> \> \> \> \> expand cluster $C$ and continue: \\
\> \> \> \> \> $C \leftarrow C \cup J$ \\
\> \> \> \> \> \textbf{return to Step 2.} \\[0.15 in]
\> \> \> \> \textbf{Else do:} \\[0.15 in]
\> \> \> \> \> store cluster $C$: \\
\> \> \> \> \> $ \mathcal{C} \leftarrow \mathcal{C} \cup \lbrace C \rbrace$\\[0.15 in]
\> \> \> \> \> eliminate unneeded seed edges and continue: \\
\> \> \> \> \> $ E_{seeds} \leftarrow E_{seeds} \setminus E(Q)$ where $Q$ is the subgraph of $ K_n$ induced by $C$\\
\> \> \> \> \> \textbf{return to Step 1.} \\
\end{tabbing}

\subsubsection{Adjustment step}
\label{adjustment}

\noindent \underline{\textbf{first iteration adjustment algorithm:}}
\begin{tabbing}
\noindent \= Input: \hspace{3 pt} \= $ \mathcal{C} = \lbrace C_1, \ldots ,C_k \rbrace$, a family of subsets $ C_i \subset V(K_n)$ where each $C_i$ is an approximation\\ 
\> \> of some cluster (in the current graph) of relatively high edgeweight density as in \\
\> \> definition \ref{DEF: density}. Observe that because it is the first iteration $ C_i \subset X$, where $X$ is the \\
\> \> original data input.\\
\noindent \= Output: \= $ \lbrace C_{\ell _ 1}, \ldots ,C_{\ell _ s} \rbrace  \subseteq \lbrace C_1, \ldots ,C_k \rbrace $, a clustering such that for any pair $ C_{\ell _ i}$ and $C_{\ell _ j}$ their \\
\> \> intersection has $ | C_{\ell _ i} \cap C_{\ell _ j} | \leq 0.5*  \mathrm{min}( |C_{\ell _ i} | , |C_{\ell _ j} |  ) $\\

\indent \= \qquad \= \qquad \= \qquad \= \qquad \= \qquad \= \qquad \= \\
\noindent Initialize: $ t \leftarrow 1$, this index tracks the current cluster.\\
\noindent \textbf{Step 1.} \\
\> \> sort the clusters in decreasing order by edgeweight density: \\
\> \> $ \mathcal{C} =  \lbrace C_{\ell _ 1} , \ldots , C_{\ell _k} \rbrace $ and induced subgraphs $ \mathcal{Q} = \lbrace  Q_{\ell _1}, \ldots , Q_{\ell _k} \rbrace $ have \\
\> \> $ \mathrm{den}(Q_{\ell _i}) \geq \mathrm{den}(Q_{\ell _{i+1}})$ for all $i$.\\[0.15 in]
\> \> initialize current cluster:\\
\> \> set $C_{\ell _ t} $ ($t$ was initialized above)\\[0.15 in]
\noindent \textbf{Step 2.} \textbf{If} all clusters have been checked, \textbf{exit adjustment step algo}, \textbf{Else do:}\\[0.15 in]
\> \> find clusters overlapping current cluster too much:\\
\> \> $ \mathcal{H} \leftarrow \lbrace C_{\ell _ z} \; | \; z>t  \; \; \text{and} \; \; | C_{\ell _ z} \cap C_{\ell _ t} | > 0.5*  \mathrm{min}( |C_{\ell _ z} | , |C_{\ell _ t} |  ) \rbrace$\\[0.15 in]
\> \> \textbf{Step 3.} \textbf{If} $ \mathcal{H} = \lbrace  \rbrace$ \textbf{do:}\\[0.15 in]
\> \> \> \> advance position and continue:\\
 \> \> \> \> $ t \leftarrow t + 1$\\
 \> \> \> \> set current cluster as $C_{\ell _ t} $\\
 \> \> \> \> \textbf{goto Step 2.}\\[0.15 in]
 \> \> \> \textbf{Else}\\
 \> \> \> \> resolve degenerate overlap by removal:\\
 \> \> \> \> $ \mathcal{C} \leftarrow \mathcal{C} \setminus \mathcal{H}$\\[0.15 in]
 \> \> \> \> advance position and continue:\\
 \> \> \> \> $ t \leftarrow t + 1$\\
 \> \> \> \> set current cluster as $C_{\ell _ t} $\\
 \> \> \> \> \textbf{goto Step 2.}\\[0.15 in]
\end{tabbing}

\noindent \underline{\textbf{subsequent iterations adjustment algorithm:}}
\begin{tabbing}
\noindent \= Input: \hspace{3 pt} \= $ \mathcal{B} = \lbrace B_1, \ldots ,B_k \rbrace$, a family of subsets $ B_i \subset X$ where each $B_i$ is an approximation\\ 
\> \> of some cluster (in the original input data) of relatively high edgeweight density \\
\> \> as in definition \ref{DEF: density}. Each $ B_i$ is obtained from a $ C_i \in \mathcal{C} $, the output of the \\
\> \> growth step, by associating members of $ C_i \subset V(K_n)$ with the objects they represent \\
\> \> in lower hierarchical levels.\\
\noindent \= Output: \= $ \lbrace B_{\ell _ 1}, \ldots ,B_{\ell _ s} \rbrace  \subseteq \lbrace B_1, \ldots ,B_k \rbrace $, a clustering such that for any pair $ B_{\ell _ i}$ and $B_{\ell _ j}$ their \\
\> \> intersection has $ | B_{\ell _ i} \cap B_{\ell _ j} | \leq 0.5*  \mathrm{min}( |B_{\ell _ i} | , |B_{\ell _ j} |  ) $\\

\indent \= \qquad \= \qquad \= \qquad \= \qquad \= \qquad \= \qquad \= \\
\noindent Initialize: $ t \leftarrow 1$, this index tracks the current cluster.\\
\noindent \textbf{Step 1.} \\
\> \> sort the clusters in decreasing order by edgeweight density: \\
\> \> $ \mathcal{B} =  \lbrace B_{\ell _ 1} , \ldots , B_{\ell _k} \rbrace $ and induced subgraphs $ \mathcal{Q} = \lbrace  Q_{\ell _1}, \ldots , Q_{\ell _k} \rbrace $ have \\
\> \> $ \mathrm{den}(Q_{\ell _i}) \geq \mathrm{den}(Q_{\ell _{i+1}})$ for all $i$. NOTE that the subgraphs $ \mathcal{Q}$ are \\
\> \> subgraphs of the original graph input to AQCM with vertices $ X$ and \\
\> \> edgeweights $S$.\\[0.15 in]
\> \> initialize current cluster:\\
\> \> set $B_{\ell _ t} $ ($t$ was initialized above)\\[0.15 in]
\noindent \textbf{Step 2.} \textbf{If} all clusters have been checked, \textbf{exit adjustment step algo}, \textbf{Else do:}\\[0.15 in]
\> \> find clusters overlapping current cluster too much:\\
\> \> $ \mathcal{H} \leftarrow \lbrace B_{\ell _ z} \; | \; z>t  \; \; \text{and} \; \; | B_{\ell _ z} \cap B_{\ell _ t} | > 0.5*  \mathrm{min}( |B_{\ell _ z} | , |B_{\ell _ t} |  ) \rbrace$\\[0.15 in]
\> \> \textbf{Step 3.} \textbf{If} $ \mathcal{H} = \lbrace  \rbrace$ \textbf{do:}\\[0.15 in]
\> \> \> \> advance position and continue:\\
 \> \> \> \> $ t \leftarrow t + 1$\\
 \> \> \> \> set current cluster as $B_{\ell _ t} $\\
 \> \> \> \> \textbf{goto Step 2.}\\[0.15 in]
 \> \> \> \textbf{Else}\\
 \> \> \> \> resolve degenerate overlap by merging:\\[0.15 in]
 \> \> \> \> \> find optimal cluster for merging with current cluster:\\
 \> \> \> \> \> $ B^{\star} \leftarrow B_{\ell_t} \cup B_x \;$, where $ B_x \in \mathcal{H}$ is the cluster such that \\
 \> \> \> \> \> $ Q^{\star}$, the subgraph induced by $ B_{\ell_t} \cup B_x$, has $ \mathrm{den}(Q^{\star})$ \\
 \> \> \> \> \> highest among all possible choices of $ B_x \in \mathcal{H}$\\[0.15 in]
 \> \> \> \> \> update the clustering: \\
 \> \> \> \> \> $ \mathcal{B} \leftarrow \mathcal{B} \setminus \lbrace B_{\ell_t} , B_x\rbrace$ \\
 \> \> \> \> \> $ \mathcal{B} \leftarrow \mathcal{B} \cup \lbrace B^{\star} \rbrace$\\[0.15 in]
 \> \> \> \> \> re-sort the clustering and continue:\\
 \> \> \> \> \> NOTE that we only need to find the correct index \\
 \> \> \> \> \> for placement of $ B^{\star}$ to preserve $ \mathrm{den}(Q_{\ell _i}) \geq \mathrm{den}(Q_{\ell _{i+1}})$ for all $i$\\
 \> \> \> \> \> \textbf{goto Step 2.}\\[0.15 in]
 
\end{tabbing}

\subsubsection{Contraction step}
\label{contraction}

\noindent \underline{\textbf{graph contraction algorithm:}}
\begin{tabbing}
\noindent \= Input: \hspace{3 pt} \= $ \lbrace C_{\ell _ 1}, \ldots ,C_{\ell _ s} \rbrace  $ the output of the adjustment step (equivalently $ \lbrace B_{\ell _ 1}, \ldots ,B_{\ell _ s} \rbrace  $ if the \\
\> \> iteration level of AQCM is 2 or higher, in this section we will simplify the notation \\
\> \> by using the symbol $C$ only). Also the graph contraction requires the original similarity \\
\> \> data $S$ that was input to AQCM in the first iteration.\\
\noindent \= Output: \= $K_s$ a complete graph on $s$ vertices where each vertex represents a cluster in $ \lbrace C_{\ell _ 1}, \ldots ,C_{\ell _ s} \rbrace  $\\
\> \> and edgeweight function $S' : E(K_s) \rightarrow \mathbb{R}^+$ where $ S'( C_{\ell _ i} , C_{\ell _ j} )$ is the average similarity \\
\> \> (according to $S$) between the data in $ C_{\ell _ i}$ and the data in $ C_{\ell _ j}$.\\ 

\indent \= \qquad \= \qquad \= \qquad \= \qquad \= \qquad \= \qquad \= \\
\noindent \textbf{Method:} \\
\> \> The definition of $ K_s$ and $S'$ are clear as described above so here we only need to define \\
\> \> the case when $ C_{\ell _ i} \cap C_{\ell _ j} $ is non-empty. The definition we state here is general in that \\
\> \> for the case where $ C_{\ell _ i} \cap C_{\ell _ j} $ \emph{is} empty, the definition is equal to the statement of the output\\
\> \> described above.\\[0.15 in]
\> \> \textbf{edge set definitions:} We need the following definitions of edge sets of the original \\
\> \> graph on the data $X$ with edge set $E$. For $ C_i$ and $ C_j$ with $ i,j \in \lbrace \ell_1 , \ldots , \ell_s \rbrace$ we have:\\[0.15 in]
\> \> \> $E_{\beta} := \lbrace uv \in E \; | \; u \in C_i \; \; \text{and} \; \; v \in C_j \; \; \text{and} \; \; u,v \notin C_i \cap C_j  \; \rbrace$\\[0.15 in]
\> \> \> $E_{\gamma} := \lbrace uv \in E \; | \; u \in C_i \; \; \text{and} \; \; u \notin C_j \; \; \text{and} \; \; v \in C_i \cap C_j  \; \rbrace$\\[0.15 in]
\> \> \> $E_{\delta} := \lbrace uv \in E \; | \; u \notin C_i \; \; \text{and} \; \; u \in C_j \; \; \text{and} \; \; v \in C_i \cap C_j  \; \rbrace$\\[0.15 in]
\> \> \> $E_{\sigma} := \lbrace uv \in E \; | \; u,v \in C_i \cap C_j  \; \rbrace$\\[0.15 in]
\> \> \textbf{edgeweight function:} For $ C_i$ and $ C_j$ with $ i,j \in \lbrace \ell_1 , \ldots , \ell_s \rbrace$ we define $ S'$:\\[0.15 in]
\> \> \> $ E_{i,j} := E_{\beta} \cup E_{\gamma} \cup E_{\delta} \cup E_{\sigma} $\\[0.15 in]
\> \> \> $ S'(C_i,C_j) := \frac{\sum_{e \in E_{i,j}} S(e) }{| E_{i,j} |}$\\[0.15 in]

\end{tabbing}

%

\section{Automatic Cluster Selection}
\label{clust_sel}
The following algorithm for automatic cluster selection is based on the principal that a ``most significant'' clustering within the many possibilities encoded by a hierarchy tree should have clusters in which the internal similarity is significantly higher than that of their parent clusters. Within the context of a hierarchy tree $T$ with parent node $x$ and child node $y$ (representing candidate clusters $ C_x$ and $C_y$ respectively) this change in cluster density between levels can be expressed as ``density drop'' $ \mathrm{den}(C_x) - \mathrm{den}(C_y)$ (see section \ref{sub_det} for definition of density). Furthermore, among possible clusterings exhibiting a significant density drop, we prefer smaller cluster size as we are interested in a larger number of clusters when such a clustering is an option. 

The algorithm featured here assigns an edgeweight to the edge $ xy \in E(T)$ that captures the properties of density drop and cluster size and then selects a clustering as an edge cut across which the average of the edgeweights is optimal among all edge cuts in the tree $T$. The algorithm for solving the optimal average weight cut problem is seen in substeps 3.1-3.4 and proof of optimality is provided in \cite{Payne2019,Payne2020edgecuts}. 
The graph theoretic operation ``edge contraction'' plays an important role (see substep 3.4 (i)) so we clarify the notation and definition here. In the directed tree $T$, the tree ``$T$ contracted by $e$'' is notated as $T/e$ and is constructed by: 1) delete the edge $e=xy$ 2) the vertices $x$ and $y$ are merged to one vertex, that is, all in edges of $x$ and out edges of $y$ are now in and out edges respectively of the new vertex.

\medskip
\noindent
{\bf Input}.
The generalized hierarchy tree $T$ (on the data set $X$) with the root vertex $v_0$,
 the size of every cluster $C$ (corresponding to each node of $T$), the similarity density $ \mathrm{den}(C)$ of every cluster $C$ (see definition \ref{DEF: density}).

\medskip
\noindent
{\bf Output}.
$ \mathcal{C} = \lbrace C_1,...,C_s \rbrace$ a clustering of the data set $X$.

\medskip \noindent
Denote the set of leafs of $T$ by $L$.

\medskip
 \noindent
{\bf Step 1.}
For each directed edge $e = xy \in E(T-L)$,
define the edge weight
\begin{equation}
w(xy)~=~\frac{| C_y|}{ \mathrm{den} (C_y)^2 - \mathrm{den} (C_x)^2 }
\label{EQ: density-drop}
\end{equation}
where $C_x$ and $C_y$ are clusters in $X$ corresponding to
 the nodes $x$ and $y$ in $T$, respectively. That is, $x$ is the parent node of the child node $y$.

\medskip
 \noindent
{\bf Step 2.}
Find a
{\bf maximum} rooted
 spanning tree $T_{max}$ of $T$ with respect to the weight $w$.
Let $v_0$ be the root of $T_{max}$.

\medskip
 \noindent
{\bf Step 3.}
In this step, we are to find an edge cut $Q$ of $T_{max}$ separating the root $v_0$ and the set $L$ of
 leaves such that
 $\frac{\sum_{e \in Q}w(e)}{|Q|}$ is
{\bf minimum}
 among all such edge cuts.

\medskip
 \noindent
{\bf Substep 3.1.}
Calculate $ \alpha_0$
$$\alpha_0 = \frac{w(E^+(v_0))}{|E^+(v_0)|}.$$

\medskip \noindent
{\bf Substep 3.2.}
For each $ e \in E(T_{max}-L)$ calculate $\lambda(e_i)$ the {\bf contractibility of the edge $e_i$} defined as
\[  \lambda(e_i) = \frac{w(E^+(e_i)) - w(e_i)}{|E^+(e_i)| - 1}
\]
and $E^+(e_i)$ is the set of all out-edges of the head of the edge $e_i$ in the directed tree $T_{max}$.

\medskip \noindent
{\bf Substep 3.3.}
Sort the edges $e_i$ of the $E(T_{max}-L)$ so that
$$\lambda(e_1) \leq \lambda(e_2) \leq ... \leq \lambda(e_m)$$

\medskip \noindent
{\bf Substep 3.4.}
If $\lambda(e_1) < \alpha_0$ then
\begin{quote}
\begin{enumerate}
\item[(i)]Denote the {\it in edge} to $e_1$ by $e^*$. Contract $T_{max} \leftarrow T_{max}/e_1$, and
\item[(ii)] update $\lambda$ value for $e^*$, or update $\alpha_0$ if $e_1$ had no {\it in edge} (it was in $E^+(v_0)$), and
\item[(iii)] repeat Substep 3.3.
\end{enumerate}
\end{quote}

\medskip \noindent
If $\lambda(e_1) \geq \alpha_0$ then go to the final step.

\medskip \noindent
{\bf FINAL STEP.}
In the resulting (smaller) tree $T_{max}$, let
 $Q=E^+(v_0)$.
The output is
$$\{ C_x \; | \;  v_0x \in Q, \; C_x \text{is the cluster corresponding to } x \text{ in } T_{max}. \}$$

\medskip \noindent
We make an important note here regarding implementation of the above cluster selection algorithm. Recall that at the first iteration of AQCM it is not guaranteed that all data have joined a cluster. Indeed, at a given level in the hierarchy tree there may be some branch that represents a still unclustered data as of that level. We find that it is useful to remove such branches from the tree $T_{max}$ before solving the optimization problem described in step 3 above. In so doing we ensure that the resulting edge cut is related to significant drops in density of clusters between hierarchical levels without the need to define such a drop in the case of unclustered data. Furthermore this heuristic would seem relevant since it is the change in density of clusters that underpins the density drop cluster selection as a strategy.

%

\section{Hierarchical Clustering in Facebook Network Data}
\label{facebook_section}

\subsection{The data set}
\label{fb_data}
In order to test the ability of AQCM method to perform real world unsupervised learning at a scale that is large relative to the expected size of clusters we chose social network data for the test data set.
The data are the Facebook friendships recorded as a ``snapshot'' of the state of the Facebook network at UC Berkeley in 2005.
This choice of data satisfies our target of ``scale relative to expected cluster size'' because of the circumstances under which the network was formed.
In 2005 Facebook was emerging as a prominent new social media platform at a time when other platforms such as Friendster and Myspace had already paved the way for social media to become a common aspect of young Americans’ internet use.
At the time Facebook required users to sign up with an academic (university, college, or high school) email address as the platform was originally designed as an internet service to facilitate social interaction on campuses and it was within these campuses that Facebook networks first formed.
These networks of student friendships represented relatively new social relationships forming around the types of social interactions fostered by campus life such as social events, fraternity/sorority life, classroom/degree related relationships.
Moreover, college students tend to develop relatively small ``close friend social circles'' which are embedded within larger social circles related to campus activity.
Hence in Facebook network data sets taken from colleges we might expect there to exist hierarchies of network community/cluster structures where on the most local level edge dense community/cluster structures are quite small relative to the network of the entire body of students on a given campus.
The UC Berkeley data set represents 22,900 individuals so we might expect there to exist thousands of small clusters of close friends, thus the data set presents the type of challenge which AQCM was designed to address in automatic cluster detection.

\subsection{Method of applying AQCM}
\label{meth_fb_AQCM}

\subsubsection{Similarity}
\label{fb_similarity}
The UC Berkeley Facebook network is not a weighted network data set. That is, for individuals $i$ and $j$ who are friends the adjacency matrix has $ A[i,j] = 1$ and if they are not friends $ A[i,j] = 0$. Thus in order to apply AQCM it is necessary to define similarity between two network nodes/vertices. For this definition we adopt the method presented in \cite{Payne2021}. Mathematical details of the method are given in the appendix
(Subsection~\ref{sim_mat}).
 The method defines for each vertex a diffusion pattern vector representing the way the vertex impacts the rest of the network in a diffusion process. Similarity between two vertices is then defined as the similarity of their diffusion vectors using ``cosine of the angle between the two vectors'', a method common in data mining \cite{SimComp2015}.

\subsubsection{Initial results and further iterative approach}
\label{fb_refinement}
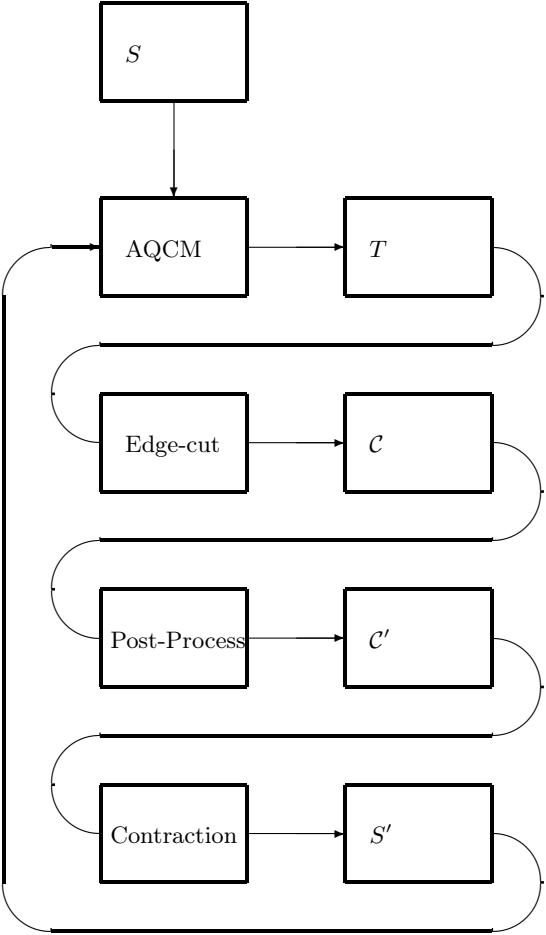
\begin{figure}[hbt!]

\setlength{\unitlength}{0.065cm}

\begin{center}

\begin{picture}(100,200)

\thicklines
\linethickness{0.5mm}

\thicklines
\linethickness{0.5mm}
\put(10,180){\line(1,0){30}}
\put(10,180){\line(0,1){20}}
\put(10,200){\line(1,0){30}}
\put(40,180){\line(0,1){20}}
\put(15,188){\small $S$}

\thinlines
 \put(25,160){\line(0,1){20}}
 \put(25,170){\vector(0,-1){10}}


\thicklines
\linethickness{0.5mm}
\put(10,140){\line(1,0){30}}
\put(10,140){\line(0,1){20}}
\put(10,160){\line(1,0){30}}
\put(40,140){\line(0,1){20}}
\put(15,148){\small AQCM}

\thinlines
 \put(40,150){\line(1,0){20}}
 \put(50,150){\vector(1,0){10}}

\thicklines
\linethickness{0.5mm}
\put(60,140){\line(1,0){30}}
\put(60,140){\line(0,1){20}}
\put(60,160){\line(1,0){30}}
\put(90,140){\line(0,1){20}}
\put(65,148){\small $T$}

\thinlines
\linethickness{0.5mm}
\put(90,140){\oval(20,20)[r]}
\put(10,130){\line(1,0){80}}
\put(10,120){\oval(20,20)[l]}


\thicklines
\linethickness{0.5mm}
\put(10,100){\line(1,0){30}}
\put(10,100){\line(0,1){20}}
\put(10,120){\line(1,0){30}}
\put(40,100){\line(0,1){20}}
\put(15,108){\small Edge-cut}

\thinlines
 \put(40,110){\line(1,0){20}}
 \put(50,110){\vector(1,0){10}}

\thicklines
\linethickness{0.5mm}
\put(60,100){\line(1,0){30}}
\put(60,100){\line(0,1){20}}
\put(60,120){\line(1,0){30}}
\put(90,100){\line(0,1){20}}
\put(65,108){\small ${\cal C}$}

\thinlines
\linethickness{0.5mm}
\put(90,100){\oval(20,20)[r]}
\put(10,90){\line(1,0){80}}
\put(10,80){\oval(20,20)[l]}


\thicklines
\linethickness{0.5mm}
\put(10,60){\line(1,0){30}}
\put(10,60){\line(0,1){20}}
\put(10,80){\line(1,0){30}}
\put(40,60){\line(0,1){20}}
\put(12,68){\small Post-Process}

\thinlines
 \put(40,70){\line(1,0){20}}
 \put(50,70){\vector(1,0){10}}

\thicklines
\linethickness{0.5mm}
\put(60,60){\line(1,0){30}}
\put(60,60){\line(0,1){20}}
\put(60,80){\line(1,0){30}}
\put(90,60){\line(0,1){20}}
\put(65,68){\small ${\cal C}'$}

\thinlines
\linethickness{0.5mm}
\put(90,60){\oval(20,20)[r]}
\put(10,50){\line(1,0){80}}
\put(10,40){\oval(20,20)[l]}


\thicklines
\linethickness{0.5mm}
\put(10,20){\line(1,0){30}}
\put(10,20){\line(0,1){20}}
\put(10,40){\line(1,0){30}}
\put(40,20){\line(0,1){20}}
\put(12,28){\small Contraction}

\thinlines
 \put(40,30){\line(1,0){20}}
 \put(50,30){\vector(1,0){10}}

\thicklines
\linethickness{0.5mm}
\put(60,20){\line(1,0){30}}
\put(60,20){\line(0,1){20}}
\put(60,40){\line(1,0){30}}
\put(90,20){\line(0,1){20}}
\put(65,28){\small $S'$}

\thinlines
\linethickness{0.5mm}
\put(90,20){\oval(20,20)[r]}
\put(0,10){\line(1,0){90}}
 \put(0,80){\oval(20,140)[l]}
  \put(0,150){\vector(1,0){10}}


\end{picture}
\end{center}
\caption{\small\it Iterative application of AQCM and edge-cut cluster detection}
\label{FIG: Hierachical tree}
\end{figure}

We applied AQCM to build a hierarchy tree from the similarity described above and then used edge-cut clustering to obtain 4,445 clusters covering approximately 15,000 of the 22,900 vertices of the Facebook network. The hierarchy tree output by AQCM has 86,430 vertices and has 15 hierarchical levels.
Clearly such a complex hierarchical clustering output presents challenges for visualization and interpretation of the clusterings it describes.
These challenges motivated the development of a further iterative process within which AQCM and edge-cut clustering serve as subroutines.
The iterative process is detailed in figure \ref{FIG: Hierachical tree}. It creates a simplified hierarchy tree based on the output of AQCM and edge-cut clustering run subsequently in a manner similar to AQCM itself.
The method also applies some simple post-processing algorithms to clustering $ \mathcal{C}$ (details provided in Appendix \ref{post_pro}) to remove multimembership and force unclustered data into clusters in a refined clustering $ \mathcal{C}'$. The purpose of post processing is so that diagrams may more easily describe the data set and validate the existence of hierarchical structures.
For the contraction step in this iterative process we use the same contraction described in section \ref{contraction}. The resulting similarity data $ S'$ is input to AQCM and the process repeats.
We were able to obtain a simplified hierarchy tree suitable for diagrams in three iterations of this process. Importantly, the run time for the iterative method was not significantly longer than simply running AQCM with edge cut clustering. This is because the large share of time is spent reducing the job from 22,900 data down to roughly 5,000 data (the size of the first clustering) at which point the job size is no longer large so the following iterations finish relatively quickly.

\newpage

\subsection{Assessment of discovered hierarchical structures}

\subsubsection{Similarity and connectivity}
\label{assess_fb}
\begin{figure}[!htb]
\centering
\includegraphics[scale=0.58,trim=25 40 0 0]{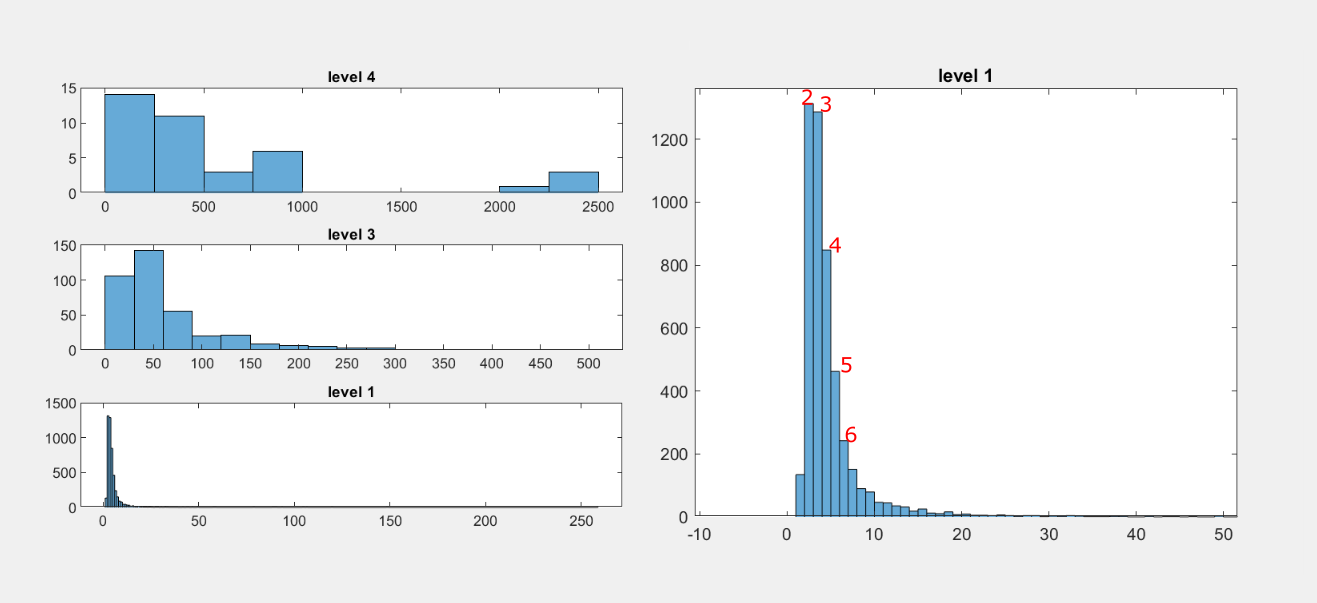}
\captionsetup{font=footnotesize}
\caption{\emph{left} histograms of cluster sizes at each of the indicated levels \emph{right} zoom in of the non-outlier cluster sizes at level 1}
\label{fb_data_fig1}
\end{figure}

\begin{figure}[!htb]
\centering
\includegraphics[scale=0.58,trim=25 40 0 0]{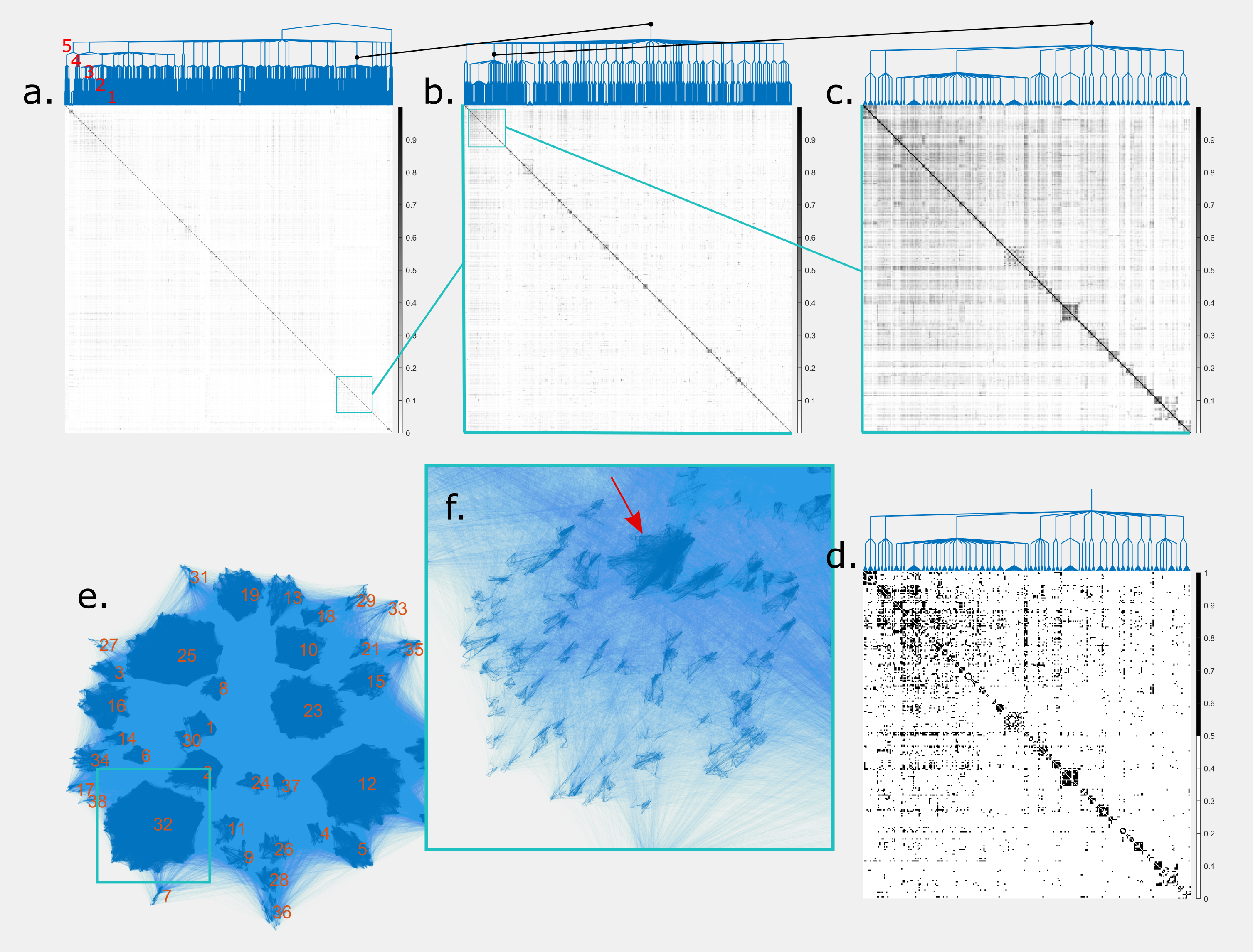}
\captionsetup{font=footnotesize}
\caption{ \emph{similarity and connectivity} \textbf{a.} the hierarchy tree output by the iterative process discussed in section \ref{fb_refinement} with five hierarchical levels labeled in red, the similarity data is shown as a matrix indexed by the tree. \textbf{b., c.} successive zoom-ins of chosen subsets as discussed in section \ref{assess_fb} \textbf{d.}  same zoom-in as in c above but the data displayed is the adjacency (connectivity) data \textbf{e.} graphical layout of the UC Berkeley Facebook network with level 4 clusters shown darker and with cluster numeric labels \textbf{f.} zoom-in of the inset blue box in e, here dark edges illustrate clusters at level 3, the red arrow indicates level 3 cluster number 23 discussed in section \ref{assess_fb}}
\label{fb_fig1}
\end{figure}

The hierarchy tree output by the process described in section \ref{fb_refinement} has five significant hierarchical levels and is pictured in figure \ref{fb_fig1}a (level numbers illustrated in red) with the similarity data described in section \ref{fb_similarity}  seen underneath as indexed by the tree.
For the purpose of demonstrating the validity of hierarchical structures discovered in the UC Berkeley network we focus on levels 1, 3, and 4 as these provide a sufficient view of the scale of the clusters at higher and lower levels.
These three levels feature clusterings of 4800, 372, and 38 clusters respectively.
Note that here we refer to non-trivial clusters, at level 1 there are 134 un-clustered points.
Some statistics on cluster size are provided for each of these three levels in table \ref{cl_stats} and histograms of cluster size for each of these levels are provided in figure \ref{fb_data_fig1}.
Level 1 has a large number of small clusters with sizes 2 and 3 accounting for 2599 of the clusters and sizes 4, 5, and 6 accounting for 1552 of the clusters.
There are 315 level 1 clusters in size range 10 through 50 and only 14 with size greater than 50 with only five of those having size greater than 100.

\begin{center}
	\begin{tabular}{| c | c | c | c | c | c |}
	\hline
	level & $\mu $ & $\sigma $ & med & min & max \\ \hline
	4 & 602.63 & 657.07 & 337.50 & 46 & 2490 \\ \hline	
	3 & 61.56 & 55.99 & 44 & 8 & 501 \\ \hline	
	1 & 4.74 & 7.2 & 3 & 2 & 259 \\ \hline
	\end{tabular}
	\captionof{table}{cluster size statistics}
	\label{cl_stats}
\end{center}

In order to provide deeper perspective on the complex structures found in the similarity data and their relationship to the social structure of the Facebook network we follow an exploratory approach along a particular branch of the hierarchy tree.
The branch represents level 4 cluster number 32 with 2490 members and is indicated with a black dot in figure \ref{fb_fig1}a while the internal similarity data of the cluster is highlighted by the blue box figure \ref{fb_fig1}a.
The internal similarity is relatively high compared with the cluster's relatively low similarity to the rest of the data and visual inspection of figure \ref{fb_fig1}a indicates the cluster is distinguished in that way within the set of level 4 clusters.
Figure \ref{fb_fig1}e shows a depiction of the connectivity of the Facebook network with level 4 clusters illustrated by darker edges and cluster numeric labels.
We computed the modularity~\cite{newman2006modularity} of each of the level 4 clusters and found cluster 32 has the highest modularity. Finally, accessing the metadata available, we note that 80\% of cluster 32 are members of the freshman class and of the freshman class 57\% are in cluster 32.
Thus we can summarize that cluster 32 is suitably interesting as it has distinctness in it's similarity and modularity properties and also correlates significantly with the social substructure induced by the freshman class. The existence of a distinct social subnetwork correlating with some active subset of the freshman class fits with our theory that the Facebook network should correspond with aspects of campus life. Indeed, we expect that freshmen seeking to establish a social life in their new environment would use Facebook friendships to facilitate social activity and thus a freshman oriented subnetwork should emerge.

Figure \ref{fb_fig1}b shows a zoom in of the similarity data inside cluster 32 with the corresponding branch of the hierarchy tree. At this zoom level the lower level small clusters of high similarity begin to become visible as blocks along the diagonal of the matrix.
Figure \ref{fb_fig1}f shows a zoom in of the area in the inset blue box in figure \ref{fb_fig1}e. In figure \ref{fb_fig1}f the dark edges illustrate the clusters at level 3.
The red arrow indicates level 3 cluster number 23. This cluster stands out as being significantly larger.
The cluster's internal similarity is highlighted by the inset blue box in figure \ref{fb_fig1}b.
Visual inspection of figure \ref{fb_fig1}b reveals the relatively higher amount of similarity inside cluster 23.
Interestingly cluster 23 also has relatively high similarity to many of the other clusters in figure \ref{fb_fig1}b,f.
This property indicates that cluster 23 might be a high centrality cluster representing a hub within the Facebook network of cluster 32.
In order to investigate this conjecture we compute the node degree (number of edges at a node) within cluster 32 for each member. For members of cluster 23 the average degree inside cluster 32 is 40 with a maximum of 184 while members of cluster 32 not in cluster 23 had average degree 21 with a maximum of 134. Furthermore, while cluster 23 members make up only 11\% of cluster 32, more than half of ``higher degree'' (degree more than 50) cluster 32 members are in cluster 23.
It is noteworthy though that cluster 23 is not entirely characterized as a group of high degree nodes. We observe that the average degree within cluster 32 is 24 and cluster 23 has 58\% with degree greater than this, hence the degree distribution skew for cluster 23 is not very large. However, cluster 23 has 23 members with degree at least 100 while members of cluster 32 not in cluster 23 account for only 4 with degree at least 100.
Hence we may characterize that cluster 23 contains almost all of the high centrality members within cluster 32 and the high centrality of a select few members of cluster 23 appears to be a key feature of the subnetwork defined by cluster 23.

Figure \ref{fb_fig1}c,d shows the zoom-ins of the internal similarity data and adjacency data respectively of cluster 23. The adjacency data shows a complex system of smaller edge-dense submodules connected together by hub-like connectivity emanating from several clusters seen at top left.
Hence the characterization of cluster 23 described above appears accurate on inspection. In the analysis below we further confirm the hub-like structures in the hierarchical subnetworks of clusters 32 and 23.

\subsubsection{Modular properties and hub-like structures}
\label{assess_fb2}
\begin{figure}[!htb]
\centering
\includegraphics[scale=0.58,trim=25 40 0 0]{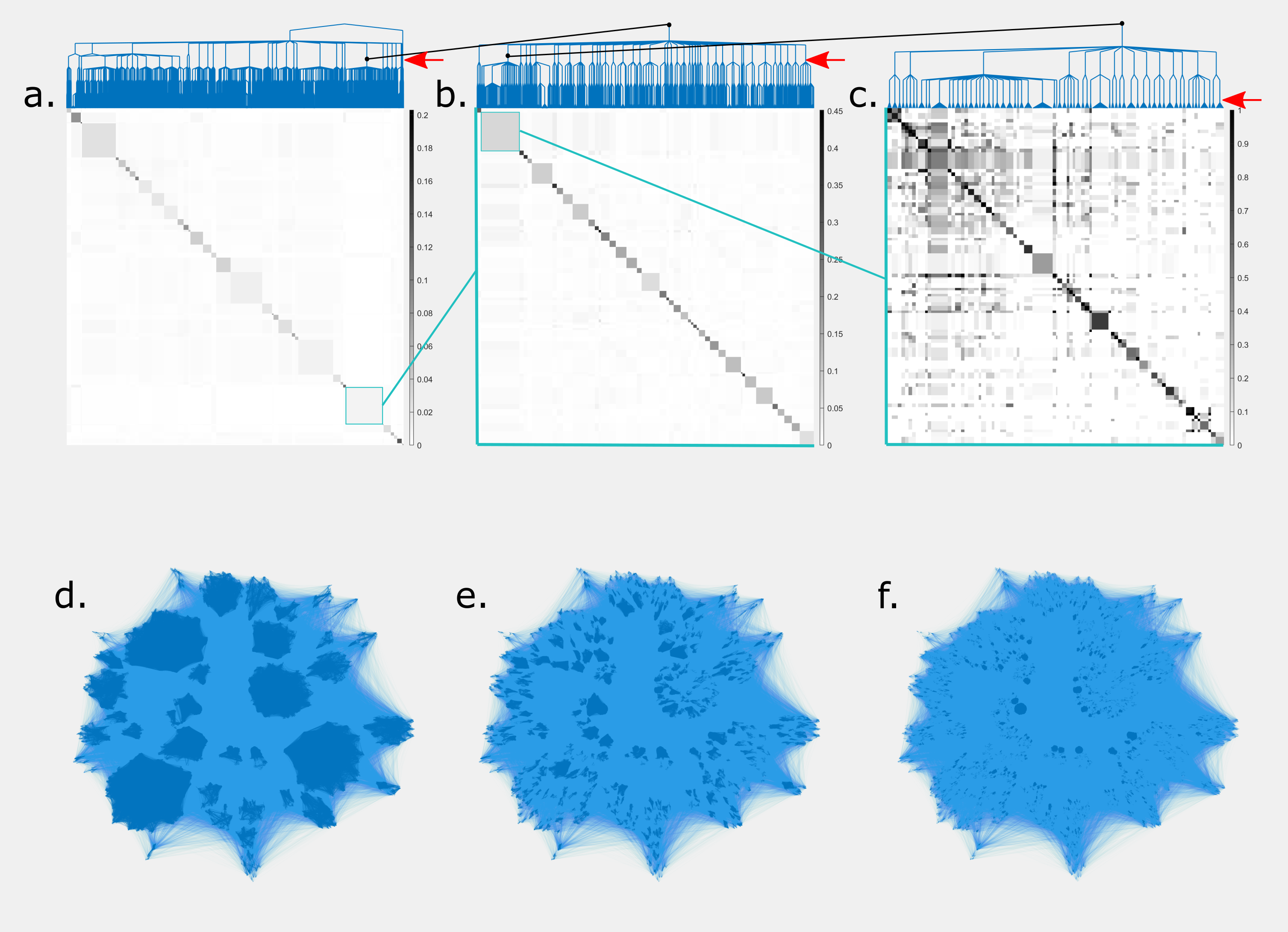}
\captionsetup{font=footnotesize}
\caption{\emph{modular properties} \textbf{a. b. c.} Shown are edge probabilities (see section \ref{assess_fb2}) within and between clusters at each of levels 4, 3, and 1 respectively. Red arrows show the hierarchical level at which edge probabilities were calculated. The zoom-ins match those in figure \ref{fb_fig1}a,b,c. \textbf{d. e. f.} Shown are graphical layouts of the connectivity of the UC Berkeley Facebook network with vertex placement matching figure \ref{fb_fig1}e. Darker edges illustrate the in-cluster edges at levels 4, 3, and 1 respectively.}
\label{fb_fig2}
\end{figure}

For the purpose of understanding the complex relationships within and between clusters we quantify the modular properties of clusters using a stochastic block model approach. For a graph $G$ with a clustering $ \mathcal{C} = \lbrace C_1,\ldots,C_s \rbrace$ of the vertex set $V$ the probability of an edge between clusters $ C_{\ell} $ and $ C_k $ is defined as the total number of such edges divided by the possible number of such edges.
If $ \ell = k$ then we count the number of edges inside $ C_{\ell}$ and divide by the possible number of such edges.
For notation we use $ P_{\ell k}$. Also note that $ P_{\ell k}$ can equivalently be referred to as edge density.
Community structures in networks are understood to be relatively edge-dense subsets of vertices separated from the rest of the network by a relatively sparse edge cut. Under this model we would say a cluster (community) $C_{\ell} $ should have $P_{\ell \ell} > P_{\ell x}$ where $ C_x$ is defined as the vertex set $V \setminus C_{\ell}$.
In this work we are interested in all the values $ P_{\ell k}$ over the clustering $ \mathcal{C}$.
For visual analysis we define a matrix $ M_{\mathcal{C}}$ based on these values.
For vertices $ v_i \in C_{\ell}$ and $ v_j \in C_k$ we assign $ M_\mathcal{C} [i,j] = P_{\ell k}$.
We computed the matrix $M_{\mathcal{C}}$ for each of the three clustering levels 1, 3, and 4.
$M_{\mathcal{C}}$ for level 4 is seen in figure \ref{fb_fig2}a as indexed by the hierarchy tree above it.
The red arrow indicates level 4.
Figures \ref{fb_fig2}b,c show $M_{\mathcal{C}}$ for levels 3 and 1 respectively and zoomed in the same way as in figures \ref{fb_fig1}b,c.

In order to assess the distinctness of the clusters in terms of their modular properties as described above we may consider, for a cluster $C_{\ell} $, the ratios of the form $ (P_{\ell \ell} - P_{\ell k}) / P_{\ell \ell}$ over all choices of $k \neq \ell$. For a cluster $C_{\ell} $ we will notate this ratio as $ \Delta_{\ell}( k )$. Furthermore it is useful to consider, for a cluster $C_{\ell} $, the minimum over $ k$ of $ \Delta_{\ell}( k )$. For this minimum we use the notation $ \delta_{\ell}( k )$.

Using the framework described above we first observe that for the level 4 clustering $ \Delta_{\ell}( k )$ is positive for each $ C_{\ell}$ with an average of 94\%. The average of $\delta_{\ell}( k )$ over all clusters is 83.8\% with standard deviation 11.8\%. Also, the minimum value of $ \delta_{\ell}( k )$ observed is 51.9\% indicating that even in the case where two clusters are not very distinctly separated there is still a significantly lower probability of between cluster edges. Hence the level 4 clustering exhibits significant community structure in that the difference in internal and between cluster edge probability is statistically significant relative to cluster edge density. In particular we have for cluster 32 the average over $k$ of $ \Delta_{32}( k )$ is 88.9\%. For 86.5\% of the $k$ values $ \Delta_{32}( k )$ is at least 75\% validating that cluster 32 is a strongly separated community, a property that agrees with its high modularity as mentioned in section \ref{assess_fb}. We may also observe that visual inspection of figure \ref{fb_fig2}a reveals that the edge probabilities between cluster 32 and most other clusters in level 4 is significantly lower than between cluster probabilities among other clusters, further demonstrating that cluster 32 stands out as a particularly well separated subnetwork.

Repeating the analytic inquiry above for the clustering at level 3 we find the following . $ \Delta_{\ell}( k )$ is positive for each $ C_{\ell}$ with an average of 98.9\%. The average of $\delta_{\ell}( k )$ over all clusters is 85.4\% with standard deviation 7.9\%. Also, the minimum value of $ \delta_{\ell}( k )$ observed is 38\% indicating that in some cases cluster separation is not as distinct and may be on account of a cluster group forming a higher level cluster. Hence the level 3 clustering exhibits community structure distinctness similar to that at level 4. Indeed, if we explore the internal connectivity of level 4 cluster 32 as an example, the level 3 cluster to cluster edge probabilities seen in figure \ref{fb_fig2}b clearly show the relatively edge dense level 3 clusters interconnected by relatively sparse connectivity.

Another feature that stands out in visual inspection of figure \ref{fb_fig2}b is the apparently higher edge probability of edges to cluster 23 (blue inset box). While the edge probability from any subcluster of 32 to subcluster 23 averages 0.9\%, the expected value of such averages for other subclusters seen in figure \ref{fb_fig2}b is just 0.47\% and for 90\% of subclusters of 32 that average is less than 0.7\%. Cluster 23 of course has the highest edge probability average hence, in the sense of the statistics mentioned here, we may confirm our conjecture made in section \ref{assess_fb} that subcluster 23 is a hub within cluster 32.

Repeating the analytic inquiry above for the level 1 clustering is more challenging on account that 1039 of the 4800 level 1 clusters have the property $ P_{\ell \ell} < P_{\ell k}$ for at least one value of $k$. Clusters such as this can arise when common neighborhood properties are more relevant for a set of vertices than direct connections. Consider for example ten Facebook members who are all friends with each other. Suppose there are three additional members who are not friends with each other but who are each friends with all of the ten. In this case the three would form a natural cluster of similarity but with no internal edge density in their group. In order to describe the edge dense properties of level 1 clusters we focus the statistics on those with $ P_{\ell \ell} > P_{\ell k}$. There are 3087 such clusters. Among those the average of $ \Delta_{\ell}( k )$ is 99.7\% and the average of $\delta_{\ell}( k )$ over those clusters is 40.6\%. Hence 64\% of level 1 clusters are simply described by the edge-dense community model.

Inspecting figure \ref{fb_fig2}c it is clear that one of the most noticeable features is the apparently hub structure formed by the prominent cluster top left. This cluster is number 1123 of the level 1 clustering. The average of $ P_{1123,k}$ over clusters $k$ featured in figure \ref{fb_fig2}c is 17.34\% which puts cluster in the top 8.9\% of subclusters of cluster 23 in terms of connectivity to other clusters. At 14 members, cluster 1123 is significantly larger that other cluster 23 subclusters with hub-like connectivity, hence we may think of it as a prominent feature of the subnetwork of level 3 cluster 23 as conjectured in section \ref{assess_fb}.

%

\section{Appendix}
\label{App}

\subsection{Additional graph theory terminology}
\label{app_graph}

For a vertex $v$ and an edge $e$ (in an undirected graph), we say $e$ is \emph{incident} to $v$ if $ e = uv$ for some vertex $u$. The \emph{degree} of a vertex $v$ is the number of edges incident to $v$ and is notated $ d(v)$. If $G$ is a weighted graph then degree is generalized as $ d(v) = \sum_{u \in V} w(uv)$, where if $ uv \notin E$ then $ w(uv) = 0$. If $v$ is a vertex of a directed graph we separately count the \emph{out-degree} (the number of edges with $v$ as a tail) and the \emph{in-degree} (the number of edges with $v$ as a head), and they are notated $ d^+(v)$ and $ d^-(v)$ respectively. Generalization of weighted in and out degree follow the same intuition as for the undirected case.

A \emph{path} between two vertices $ u,v \in V$ is a sequence $ u = u_1 u_2 ...u_k = v$ such that for each consecutive pair $ i,i+1$ we have, $ u_i u_{i+1} \in E$ and each vertex appears only once in the sequence. For two specific vertices $u$ and $v$ a path between them is often called a $u,v-$path. A \emph{closed path} is a path where the first and last vertex are the same and is often called a \emph{cycle}. We say a graph is \emph{connected} when for any pair of vertices $ u,v \in V$ there is a $u,v-$path. A set of edges $ H \subset E$ is called an \emph{edge cut} when the removal of the edges $H$ causes the graph to become disconnected. In many graph theoretic works it is important to restrict study to edge cuts that split a graph into a set of \emph{non-trivial} components $ \lbrace Q_1,...,Q_k \rbrace$. A non-trivial component $Q$ of a graph $G$ is a \emph{subgraph} of $G$ with $ V(Q) \subsetneq V(G)$ and $ E(Q) \subsetneq E(G)$ and $ | V(Q) | >1$ and $ Q$ is connected and importantly for $ u,v \in V(Q)$ with $ uv \in E(G)$ then $ uv \in E(Q)$. Edge cuts that split a graph $G$ into non-trivial components are called non-trivial edge cuts, but in context it is often understood, in a given work, that the term \emph{edge cut} refers to those that are non-trivial.

A minimal connected graph on a vertex set $V$ with $ |V| = n$ is any graph such that no edge can be removed without disconnecting the graph. It is easy to see that any such graph has no cycle and easy to show that any such graph has $ n-1$ edges. A minimal connected graph is called a \emph{tree} and often such a graph is notated $T$ instead of $G$. An obvious property of any tree $T$ is that for any pair of vertices $u$ and $v$ there is exactly one $u,v-$path.

In the study of directed graphs, an important type of tree is the \emph{rooted tree}. A rooted tree is a directed tree with exactly one vertex $v$ with $ d^-(v) = 0$ (this vertex is called the \emph{root}) and set of vertices $ \lbrace u_1,...,u_k \rbrace$ with $ d^+(u_i) = 0$ for each (these vertices are called \emph{leaves}). Stated another way, for a rooted tree $T$, there is one vertex (root) that is the \emph{head} of no edge and some set of vertices (leaves) that are each the tail of no edge. It is easy to show that for a root vertex $v$ and any non-root vertex $u$ there is a $v,u-$dipath (directed path), hence a rooted tree can be thought of as having a ``layered'' structure in that nodes can be organized into groups by their distance from the root along root-to-leaf paths.


\subsection{The Similarity Matrix, $S$}
\label{sim_mat}
We give here a brief summary of the construction of the matrix $S$, 
 that we used in our analysis of the FaceBook network data.  Detailed study of this construction of similarity matrix can be seen in \cite{Payne2021}.

\noindent {\bf Input}. Given an adjacency matrix $A$ of an input graph $G$ where the $(i,j)$-entry $A[i,j]$ is the weight of the directed edge $v_i \rightarrow v_j$. Let $n=|V(G)|$. (If $G$ is undirected, $A$ is symmetric; If $G$ is unweighted, $A[i,j]=0$ or $1$.)

\medskip
\noindent {\bf Output}. A diffusion similarity matrix $S$ where the
$(i,j)$-entry $S[i,j]$
 is the similarity between
 the vertices $v_i$ and $v_j$.

\medskip
NOTE: Performing {\bf Step 1.} for any directed graph input will save computation time needed to otherwise confirm the strong-connected property for the graph.

\medskip
\noindent
{\bf Step 1.} If $G$ is strongly connected, then go to Step 2.
Otherwise, let $v_{n+1}$ be a new vertex, and
$$V(G) \leftarrow V(G) \cup \{ v_{n+1}\},$$
$$E(G) \leftarrow E(G)
 \cup \{ v_i \rightarrow v_{n+1}:~ i=1, \cdots, n \}
\cup \{ v_{n+1} \rightarrow v_i:~ i=1, \cdots, n \}$$
 and
for every $ i=1, \cdots, n$,
$$A[i,n+1]=A[n+1,i]~=~ 10\% \min \{ A[i,j]:~ A[i,j] > 0, \forall i, j = 1, \cdots, n \} $$
then go to Step 2.

\medskip
\noindent
{\bf Step 2.} Construct $W$
  where the
$(i,j)$-entry

\begin{equation}
W[i,j] :=
  \frac{A[i,j]}{\sum_{\forall \mu} A[i,\mu]}
\label{EQ: Markov matrix}
\end{equation}

\medskip
\noindent
{\bf Step 3.} Construct

\begin{equation}
  K ~ :=
 ~
\sum_{k=1}^\infty \frac{( \approx 1.63)^k}{k!} W^k
\label{EQ: heat diffusion}
\end{equation}

\medskip
\noindent
{\bf Step 4.}
If $|V(G)| = n+1$ (that is, Step 1 was taken), then deleting the $(n+1)$st-row and $(n+1)$st-column from $K$.
And go to Step 5.

If $|V(G)|=n$, then go to Step 5.

\medskip
\noindent
{\bf Step 5.}
Let
$\vec{K}_{row}(i)$ be the $i$-th-row of $K$ and
$\vec{K}_{col}(i)$ be the $i$-th-column of $K$.
Construct the output $S$ where the $(i,j)$-entry
\begin{equation}
S[i,j] ~ :=
~ \frac{ \cos(\vec{K}_{col}(i), \vec{K}_{col}(j)) ~ + ~
\cos(\vec{K}_{row}(i), \vec{K}_{row}(j))}{2}
\label{EQ: our S}
\end{equation}
\noindent where $\cos(\vec{\alpha}, \vec{\beta})$ is the cosine of the angle between vectors $\vec{\alpha}$ and $\vec{\beta}$.
\medskip
\noindent
{\bf END.}

\subsection{Post processing}
\label{post_pro}

Let $X$ be a data set and $ \mathcal{C} = \lbrace C_1,...,C_s \rbrace$ be a family of subsets of $X$, possibly with $ |C_i| = 1$ for some
$C_i$ and possibly with some pair of
subsets having non-empty intersection. Let $ S(x,y) $ be the similarity value of data $ x$ and $y $ and let the similarity density of $ C_i$, notated $\mathrm{den}(C_i )$, and the contribution of $x$ to $ C_i$, notated $\mathrm{cont}(x, C_i)$, be as defined in
Section~\ref{sub_det}. If $ |C_i | = 1$ then we shall define that $\mathrm{den}(C_i ) = 1$. The following post processing algorithms augment the clustering $ \mathcal{C} $ so as to accommodate certain properties of clustering output that may be desirable for some applications.

\subsubsection{Expanding the cluster set by including some un-clustered vertices}
The AQCM algorithm builds a hierarchical description of a data set based on preserving local similarity density. It is possible for a given data set that the output clustering of AQCM and cluster detection will leave some data points unclustered as ``outliers'' of a significantly more locally dense selected cluster. For various data analytic purposes it may be desirable to place these unclustered data into clusters. These unclustered data also may naturally form small clusters themselves if their local cluster was detected at a position in the hierarchy tree higher than a cluster detection cut for some nearby denser cluster. For these reasons we have developed the following post processing algorithm which allows unclustered data to naturally ``self-assign'' to another unclustered data or a nearby cluster based on a factor described below.

\medskip
The following definitions quantify the similarity of data $x$ to $ C_i$ as relative to other clusters and also as relative to the similarity density of $ C_i$. In this way the \emph{clustering factor} $ \phi_c(x)$ is a value that represents a best fit of data $x$ to a cluster in that data $x$ should be most similar to it's ideal cluster without reducing the cluster's similarity density too much.

\begin{definition}
\label{defmutpref}
For a given $x \in X$
 and $C_i \in \mathcal{C}$ with $ C_i \neq \{ x \}$,

 (1) The {\em individual preference} of $x$ to $C_i$.
\[ \phi_p(x,C_i )
= \frac{\mathrm{cont}(x,C_i ) }{\mathrm{max} \lbrace \mathrm{cont}(x,C_j ) \; | \; C_j \in \mathcal{C} \; with \;  C_j \neq \{ x \}  \rbrace}
\]\\

(2) The {\em community acceptance} of $x$ by $C_i$.
\[ \phi_a(x,C_i )
= \frac{\mathrm{cont}(x,C_i ) }{\mathrm{den}(C_i) }
\]\\

(3) The {\em mutual preference} between $x$ and $C_i$.
\[ \phi_m(x,C_i) = \phi_p(x,C_i) \; \mathsf{x} \; \phi_a(x,C_i)
\]

(4) The {\em clustering factor} of $x$,
$$ \phi_c(x)~=~ \max \{\phi_m(x,C_i) \; | \; C_i \in \mathcal{C} \}.$$
\end{definition}

\subsubsection{Expansion algorithm}
 \label{sub22}

\medskip \noindent
{\bf Input:}
 $\mathcal{C}=\{ C_1,...,C_s \}$ a clustering of a data set $X$ with $|C_i| \geq 2 \; \; \forall i$ , and a threshold parameter $ \rho $.

\medskip \noindent
{\bf Output:}
  $\mathcal{C}' =\{ C_1,...,C_t \} $, with $ t \geq s$ and $ \mathcal{C}'$ covers more of the data $X$ than $ \mathcal{C}$.

\medskip \noindent
{\bf Step 1:}
List the unclustered data: $X' \leftarrow \{ x \in X \; | \; x \notin C_i ~\mbox{for~any}~
C_i  \in {\cal C} \}$.

\medskip \noindent
{\bf Step 2:}
Update ${\cal C}$ as follows: for each $ x \in X'$ create a cluster $ C = \{ x \}$ and store the cluster $ \mathcal{C} \leftarrow \mathcal{C} \cup \{ C \}$.

\medskip
\noindent
{\bf Step 3:}
For each $x \in X' $ and every $C_i \in {\cal C} $,
calculate $\phi_c(x) $ and $\phi_m(x, C_i)$.

\medskip
\noindent
{\bf Step 4:}
For each $ x \in X'$, if there is a unique cluster $C_i$ for which
$\phi_c(x) = \phi_m(x,C_i)$, and
  $\phi_c(x) \geq \rho$, then
  add $x$ to $C_i$.

\medskip
\noindent
{\bf Step 5:}
From $\mathcal{C}$,
 remove $C_i$ if $|C_i| \leq 1$, and
 delete any duplicate cluster. $ \mathcal{C}' \leftarrow \mathcal{C}$ and output $ \mathcal{C}'$.

\medskip
\noindent
{\bf END}

\subsubsection{Elimination of multi-membership property}
The growth subroutine of AQCM allows clusters to include optimally similar points regardless of whether those points have already joined another cluster to which they are also optimally similar. Thus points representing the boundary of clusters may be ``multi-members''. For some analytic purposes however it may be desirable to enforce a strict partitioning of a data set into clusters. For this reason we provide the following simple method to require multi-member data points to ``choose'' one cluster in which to remain a member.

\medskip
We may notate the multi-member set precisely. Let $ X'' \subset X$ such that for $ x \in X''$  $ \exists \mathcal{H}_x \subset \mathcal{C}$ where $ x \in C_i \; \forall C_i \in \mathcal{H}_x$ and $ | \mathcal{H}_x | \geq 2$.

\begin{definition}
\label{defmultimem}
For a given $x \in X''$
 and $C_i \in \mathcal{C}$

(1) The contribution of $x$ \emph{to the core of} $ C_i$.
\[ \varphi (x,C_i )
= \mathrm{cont}( x , C_i \setminus X'' )
\]

(2) The \emph{core factor} of $x$.
\[ \varphi_c(x)
= \max \lbrace \varphi(x,C_i) \; | \; C_i \in \mathcal{C}  \rbrace
\]

\end{definition}

\subsubsection{Multi-membership elimination algorithm}
 \label{elimmulti}

 \medskip \noindent
{\bf Input:}
 $\mathcal{C}=\{ C_1,...,C_s \}$ a clustering of a data set $X$ and some pairs $ C_i,C_j$ have $ C_i \cap C_j \neq \emptyset$

\medskip \noindent
{\bf Output:}
  $\mathcal{C}' =\{ C_1,...,C_t \} $, with the number of multi-member data reduced.

\medskip \noindent
{\bf Step 1:}
List the multi-member data: $X'' \leftarrow \text{multi-member data points as described above}$

\medskip
\noindent
{\bf Step 2:}
For each $x \in X'' $ and every $C_i \in {\cal C} $,
calculate $\varphi_c(x) $ and $\varphi(x, C_i)$.

\medskip
\noindent
{\bf Step 3:}
For $ x \in X''$, if there is a unique cluster $ C_i$ for which $ \varphi_c(x) = \varphi(x,C_i)$ then remove $x$ from all clusters except $ C_i$.

\medskip
\noindent
{\bf Step 4:}
From $\mathcal{C}$,
 remove $C_i$ if $|C_i| \leq 1$, and
 delete any duplicate cluster. $ \mathcal{C}' \leftarrow \mathcal{C}$ and output $ \mathcal{C}'$.

\medskip
\noindent
{\bf END}

\bibliography{AQCMRefs-20210305-CQ-V5}{}
\bibliographystyle{plain}

\end{document}